\documentclass[10pt,journal,compsoc]{./IEEEtran}

\usepackage[utf8]{inputenc} % allow utf-8 input
\usepackage[T1]{fontenc}    % use 8-bit T1 fonts
\usepackage{hyperref}       % hyperlinks
\usepackage{url}            % simple URL typesetting
\usepackage{booktabs}       % professional-quality tables
\usepackage{amsfonts}       % blackboard math symbols
\usepackage{nicefrac}       % compact symbols for 1/2, etc.
\usepackage{microtype}      % microtypography
\usepackage{color}
\usepackage{graphicx}
\usepackage{float}
\usepackage{amsmath,amssymb}
\usepackage{makecell}
\usepackage{subfig}
\usepackage{xcolor}

\begin{document}
\title{Anomaly Detection in Event-triggered Traffic Time Series via Similarity Learning}
\author{Shaoyu~Dou,
        Kai~Yang,~\IEEEmembership{Senior~Member,~IEEE},
        Yang Jiao,
        Chengbo Qiu,
        and Kui Ren, ~\IEEEmembership{Fellow,~IEEE} % <-this % stops a space
\IEEEcompsocitemizethanks{
  \IEEEcompsocthanksitem S. Dou, K. Yang, Y. Jiao and C. Qiu are with the Department of Computer Science and Technology, Tongji University, Shanghai 201800, China. Kui Ren is with Zhejiang University, China.
  E-mail: kaiyang@tongji.edu.cn \protect\\

  \IEEEcompsocthanksitem This work was supported in part by the National Natural Science Foundation of China under Grant 12371519, 61771013 and 62032021; in part by the Fundamental Research Funds for the Central Universities of China; in part by the Fundamental Research Funds of Shanghai Jiading District. \protect\\
}% <-this % stops an unwanted space

\thanks{Manuscript received ***.}}
\markboth{Journal of \LaTeX\ Class Files,~Vol.~14, No.~8, August~2015}%
{Shell \MakeLowercase{\textit{et al.}}: Bare Demo of IEEEtran.cls for Computer Society Journals}

\IEEEtitleabstractindextext{%
\begin{abstract}
Time series analysis has achieved great success in cyber security such as intrusion detection and device identification. Learning similarities among multiple time series is a crucial problem since it serves as the foundation for downstream analysis. Due to the complex temporal dynamics of the event-triggered time series, it often remains unclear which similarity metric is appropriate for security-related tasks, such as anomaly detection and clustering. The overarching goal of this paper is to develop an unsupervised learning framework that is capable of learning similarities among a set of event-triggered time series. From the machine learning vantage point, the proposed framework harnesses the power of both hierarchical multi-resolution sequential autoencoders and the Gaussian Mixture Model (GMM) to effectively learn the low-dimensional representations from the time series. Finally, the obtained similarity measure can be easily visualized for the explanation. The proposed framework aspires to offer a stepping stone that gives rise to a systematic approach to model and learn similarities among a multitude of event-triggered time series. Through extensive qualitative and quantitative experiments, it is revealed that the proposed method outperforms state-of-the-art methods considerably.
\end{abstract}

}

\maketitle

\IEEEdisplaynontitleabstractindextext

\IEEEpeerreviewmaketitle

\IEEEraisesectionheading{\section{Introduction}\label{sec:introduction}}

\IEEEPARstart{T}{ime} series anomaly detection is crucial in various domains, including cyber-security and astronomy. A central challenge in time series anomaly detection is measuring the \textit{similarity} between two time series, which is essential for comparing samples and differentiating abnormal from normal. Practically, computing a suitable similarity metric lies at the heart of numerous machine learning tasks, including clustering and supervised classification.
The motivation for learning the similarity of \textit{event-triggered time series} in this paper is to detect anomalous or malicious behavior of devices or software via their traffic. For instance, a hacked malicious healthcare IoT device may send sensitive personal information to the Internet, which compromises users' privacy and requires immediate attention \cite{ren2019information}.
More generally, machine learning applied to time series, including applications like network device behavior detection and IoT detection, is anticipated to be pivotal in various emerging applications \cite{DNA-2017,InsideAD-2021,trafficAD-2011,trafficAD2018}. This motivates the extension of similarity learning beyond just anomaly detection to a broader range of machine learning tasks, such as clustering.

\begin{figure}[!tbp]
  \centering
  \includegraphics[width=\columnwidth]{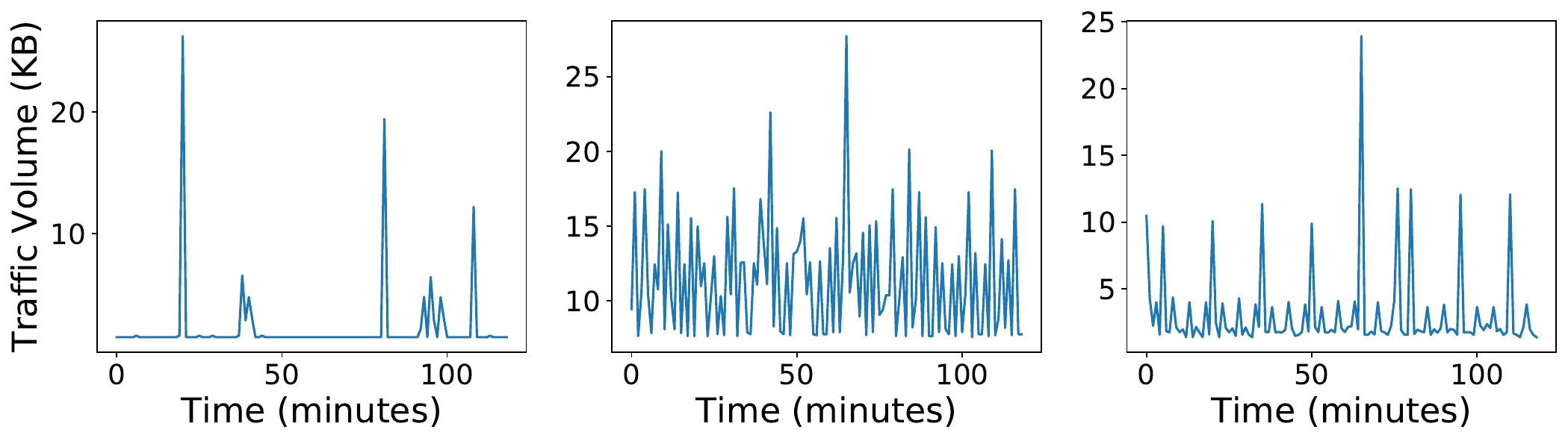}
  \caption{\textcolor{black}{Representative event-triggered traffic time series from the UNSW-IoT dataset.}}
  \label{fig:example}
\end{figure}

The past decade has witnessed a proliferation of event-triggered sensors or software-generated time series data, where events refer to human intervention or programmed machine activity.
Such events will trigger the working state transition of the program, resulting in heterogeneous dynamics of the traffic time series. Please refer to Section \ref{sec:event_triggered} for an informal definition of event-triggered time series. 
A key challenge in analyzing such event-triggered (traffic) time series is that it often contains temporal event sequences that are sporadic or highly heterogeneous as shown in Figure \ref{fig:example}. It may contain a few short traffic bursts and a long sleep time with no data transmission, as shown in the first sub-figure, or seems to be the ``superposition'' of multiple time series, as shown in the second and third sub-figures. The unique pattern makes this type of time series extremely heterogeneous and exhibits both long-term and short-term temporal dependencies which render the traditional machine learning algorithms not directly applicable. Apart from the heterogeneity, other challenges include: 
1) Since the labeling process is often very expensive, there are rarely sufficient and accurate labels for similarity learning. 
2) To achieve the best performance, similarity learning needs to be tuned for a particular task. 
3) Event-triggered sensors or software-generated traffic time series data often vary in time granularity. 
4) In many applications, we need to not only compute the similarity between two unlabeled time series but also provide insight into the mechanism so that domain experts can understand the similarity metric. 

The above challenges give rise to the following questions.

\emph{How can we design an unsupervised machine learning approach to learn the similarity between two event-triggered time series?}

The traditional sequential autoencoder, such as GRU or LSTM-based autoencoder, encounters the problem of error accumulation during autoregressive decoding \cite{bengio2015scheduled,lamb2016professor}, leading to suboptimal reconstruction and representation quality, particularly for non-smooth time series with high temporal dynamics, as shown in Figure \ref{fig:example}.
In this paper, we present ET-Net, a bagging model designed to effectively model the temporal dynamics of \textbf{E}vent-\textbf{T}riggered time series. To address the challenges mentioned above, we propose two ensemble components that adopt different autoencoders, which are named W compression network and D compression network and inspired by multi-task and multi-resolution learning, respectively. The proposed autoencoders learn robust representations and form the basis of similarity learning. Each ensemble component then utilizes a statistical Gaussian Mixture Model (GMM) to measure similarities among a collection of time series on the learned representations. The overall ET-Net provides visual outcomes to aid human understanding. 
More specifically, we have made the following contributions:

\begin{itemize}
\item \textbf{Unsupervised similarity learning:} ET-Net is completely unsupervised and can learn similarities among unlabeled event-triggered time series. In addition, similarity learning can be tuned for a particular task to optimize the detection performance.
\item \textbf{Visualization and interpretability:} ET-Net generates a semantically meaningful latent space that can be visualized to explain the learned model. In addition, some classification rules can be drawn from the visualization of the original space to help human experts understand the data.
\item \textbf{Effectiveness:} ET-Net exhibits strong empirical performance for downstream analysis such as anomaly detection and clustering than other competing state-of-the-art methods over real-world datasets. It remains effective even when the training data is contaminated by anomalies or noises that are common in time series analysis \cite{wang2013effectiveness}. In addition, the proposed model still has competitive performance on testing data with different time granularity \cite{eibl2014influence} and traffic disturbances without retraining.
\end{itemize}

%Bitnet is \textbf{open sourced} on \url{https://github.com/KMdsy/Bitnet}.

% --------------------------------------------------------------------------------------

\section{Related Work}
\subsection{Similarity Learning}
We summarize related work on similarity learning that has received significant attention over the past decade, including non-parametric methods and parametric methods based on deep neural networks.

There exists a large body of work on non-parametric methods for time series similarity learning, including Euclidean Distance (ED)\cite{keogh2002need}, Editing Distance on Real sequences (EDR)\cite{chen2005robust}, and Dynamic Time Warping (DTW)\cite{sakoe1978dynamic}. Euclidean distance, while the most widely used distance metric, often yields poor performance for time series similarity learning because it is sensitive to anomalies, noise, warping and contamination \cite{toller2019formally}. 
To address this issue, one may employ techniques such as Edit Distance with Real Penalty (EDR) and Dynamic Time Warping (DTW) to handle sequences of unequal length. However, EDR is calculated based on local procedures and treats all change operations equally, making it sensitive to noise and irregular sampling rates. Thus, even minor deviations in a group of data points within the time series can cause EDR to produce large values, compromising its effectiveness.
DTW is a time series similarity metric that has been widely studied and used in recent years. It aims to calculate an optimal matching between two time series and has proved to be capable of providing strong baseline performance in many machine learning tasks such as classification and clustering. In addition, there exists a lot of work dedicated to improving the performance of DTW \cite{silva2016effect} or combining DTW with deep learning methods \cite{cuturi2017soft, cai2019dtwnet}.

The majority of deep learning-based methods for time series similarity learning use a Recurrent Neural Network (RNN) or Convolutional Neural Network (CNN) to model the temporal dynamics and convert them into low-dimensional representations. Similarity or distance metrics in this low-dimensional latent space are expected to reflect the semantic relationship between time series. \cite{mathew2019warping} proposes a structure called WaRTEm to generate time series embedding that exhibits resilience to warping. \cite{ma2019learning} proposes a model named DTCR that integrates the seq2seq model and the K-means objective to generate latent space representations that are better suited for clustering. Autowarp proposed in  \cite{abid2018learning} obtains a vector embedding through the sequence autoencoder, which helps to guide the optimization of a warping metric.

% ------------------

\subsection{Anomaly Detection}
Time series anomaly detection can be broadly categorized as either supervised or unsupervised methods. Supervised anomaly detection techniques such as Support-Vector Machine (SVM) \cite{cortes1995support} and Random Forest (RF) \cite{breiman2001random} require a large amount of accurate labels to achieve optimal performance. However, in practice, imbalanced training data and inaccurate labeling can lead to performance degradation. Additionally, in many cases, it is not possible to obtain labels for anomaly detection, making the supervised approach completely inapplicable.
Unsupervised anomaly detection, which includes classification-based methods \cite{scholkopf2000support} and density-based methods \cite{breunig2000lof,leung2005unsupervised}, detects anomalies by training a one-class classifier on normal data points or by performing density estimation. Although these methods can achieve satisfactory anomaly detection accuracy, they are often ineffective when dealing with high-dimensional time series data.

Recently, reconstruction-based deep learning methods have emerged as a new means for unsupervised anomaly detection in high-dimensional data. These methods assume that the reconstructions of low-dimensional projections of anomalies will deviate greatly from the original samples, and use the reconstruction error to detect anomalies \cite{baldi2012autoencoders, zhou2019beatgan}. However, vanilla reconstruction-based methods are often limited because they only conduct anomaly detection based on reconstruction error.
Hybrid methods have been developed to harness the power of both autoencoders and model-driven approaches. This combined approach compresses the original data into a latent space and then performs model estimation in that space. For instance, \cite{ma2019learning} uses density-based K-means clustering in the latent space, while \cite{shen2021time} and \cite{zong2018deep} model the latent space from the perspective of probability by leveraging normal distribution and Gaussian Mixture Model (GMM), respectively. However, the assumption that the reconstruction error follows a Gaussian distribution in \cite{shen2021time} is not applicable to highly heterogeneous event-triggered time series. This is because different networks have varying reconstruction performances on heterogeneous data.
Additionally, \cite{zhaoae} learns a normalized flow in the latent space.

It is worth noting that some recent studies have explored hybrid methods or ensembled autoencoders which are also adopted in proposed ET-Net. For instance,  \cite{luo2016detection} proposed a two-stage approach that trains an SAE-based representation network with GMM, which can lead to suboptimal results due to the separate optimization. In contrast, ET-Net adopts an end-to-end approach to jointly optimize the representation learning and density estimation. As a result, the representations are constrained by GMM, leading to a visually interpretable latent space. Another example is \cite{yang2019deep}, which proposed learning graph representations with a node similarity measure based on topology as supervision, but this type of supervised information is not available for time series since there is no prior knowledge that represents the similarity among a set of time series. In \cite{an2022ensemble}, sample representations are generated using a homogeneous multichannel autoencoder, and the network is optimized in a supervised manner using domain labels. In contrast, ET-Net is a completely unsupervised architecture that learns the similarity of event-triggered time series without relying on any prior knowledge or labels.

% ------------------

\subsection{Visual Interpretability}

Visual interpretability is often seen as the first step in explaining deep neural networks. It can be used to explain the inherent mechanism of the neural network \cite{ma2019learning,zong2018deep,zeiler2014visualizing}, and can also be used to trace which training samples or features significantly affect the output of the neural network, i.e., attributing the output of neural network to a set of features or samples.
In general, the attribution methods can be roughly categorized into two groups, i.e., feature-based and example-based. The feature-based attribution methods compute the contribution of each input feature to the model output and visualize the result through a heat map superimposed on the input sample. 
LIME \cite{ribeiro2016should} optimizes a white-box model to locally approximate the output of the given neural network, and then determine the contribution of each feature by analyzing the learned white-box model.
SHAP \cite{lundberg2017unified} calculates the contribution of each feature based on game theory.
\cite{dabkowski2017real} optimizes a masking model to identify the input features that most influence the decision of the classifier.
IntegratedGrads \cite{sundararajan2017axiomatic} proposes to use the integrated gradients of the feature as the importance score of the feature.
However, such methods often struggle to generate convincing attribution results on time series data, because the model may highlight features that the model considers important but not to human experts \cite{jeyakumar2020can}.

Example-based attribution methods visualize a set of training samples or prototypes to explain the output of the network. \cite{koh2017understanding} utilizes the influence function to determine which training samples play a decisive role in the model prediction for a given sample. \cite{papernot2018deep} proposes that the K nearest neighbors of a given sample in the feature space are the training samples that contribute the most to the network output. While extensive efforts have been undertaken for the example-based attribution method, its application to time series data is still an under-explored area. 

ET-Net adopts an example-based visually interpretable approach for several reasons. Firstly, feature-based approaches such as LIME and SHAP attribute the model output to a set of data points in a time series. However, in event-triggered time series, each data point is aggregated from the behavior of multiple events, thus analyzing only a single sample, and attributing output to data points does not explore the specific event that caused the anomaly, nor help humans understand how to distinguish normal and abnormal samples.
In contrast, the example-based method provides a group of similar abnormal samples and a set of normal samples for comparison when analyzing a given abnormal sample. Although this cannot directly indicate the specific reason for the anomaly, humans can obtain knowledge to distinguish normal and abnormal samples by comparing and summarizing these samples, as discussed in Section \ref{sec:inter}.
Moreover, while some methods, such as the counterfactual sample generation-based method \cite{chen2019looks}, add perturbations to the original sample to generate samples that would alter the model prediction, this can be problematic in time series analysis scenarios. Specifically, any artificial perturbation on the time series can cause difficulties understanding the time series. Furthermore, interpretation methods designed for specific network structures, such as CNN \cite{guidotti2020black}, can be challenging to extend to the task of time series analysis.
To avoid the drawbacks mentioned above, ET-Net employs a similar approach as \cite{papernot2018deep}, but instead outputs normal and abnormal samples for a given anomaly sample in an unsupervised setting. This approach helps humans to better understand the anomaly and may summarize knowledge from that.

% 本文中，我们基于例子的视觉可解释方法，是出于以下原因：feature-based方法，如LIME、SHAP等，会将模型的输出归因于时间序列的某个某个时间点，但在事件驱动时间序列中，每个时间点的值均由多个事件的行为汇聚而成，所以归因于单个时间点既不能探究导致异常的具体事件，只分析单个样本的也不能帮助人类理解如何区分正异常。
% 而example-based方法在分析给定异常样本时，能够同时提供被同样认为是异常的一组相似样本以及作为对比的正常样本，这虽然不能直接的指明异常发生的具体原因，但人类可以通过在这些样本上做比较、归纳总结，来获得区分正异常的知识，如section 5中归纳的那样。
% 一些基于反事实样本生成的方法[35]，通过在原始样本上增加扰动以生成会改变模型预测样本。这在时间序列分析场景下会带来困扰，具体的，时间序列不同于图像，任何一个人工合成的反事实的样本可以被人类轻易的判断为与某个类别的样本相似，从而理解给定样本被分类为异常的原因，任何时间序列上的人为扰动都会对理解是时间序列带来困难。于此同时，与一些专为特定网络结构设计的解释方法，如69，也难以拓展到时间序列分析任务中。
% ET-Net采用了类似\cite{papernot2018deep}的思路避免了生成样本带来的，但在非监督的设定下为给定异常样本输出了正异常例子用于帮助人类理解异常。

In this paper, we propose an end-to-end general framework for learning the similarity between event-triggered time series in a fully unsupervised manner. The resulting outcomes can be easily visualized in latent space for human comprehension. We also use the example-based attribution method to explain the model decisions from the original space. Moreover, the proposed model learns the vector embeddings in the latent space by taking into account the machine learning task under investigation. Finally, the probabilistic GMM model offers probabilistic measurement and is more flexible than the K-means clustering method. We summarize the unique features of our model and compare it with other state-of-the-art approaches in Table \ref{table:related}. In particular, the GMM adopted in this framework adopts mixed membership and is much more flexible in terms of cluster covariance than the hard assignment approach such as the K-means clustering method \cite{ma2019learning}. The proposed task-aware approach can learn the vector embeddings that are tailored for a particular machine learning task, so the detection performance can be improved. As evident from Table \ref{table:related}, only ET-Net meets all the desired requirements.

\begin{table*}[!tbp]
  \caption{Comparison of related work}
  \label{table:related}
  \centering
  \scalebox{1}{
  \begin{tabular}{@{}lccccccc@{}}
  \toprule
                            & ED          & DTW         & EDR           & WaRTEm \cite{mathew2019warping} & DTCR \cite{ma2019learning}  & Autowarp \cite{abid2018learning}  & ET-Net     \\ \midrule
  %Interpretability          & \checkmark  & \checkmark  & \checkmark    & \checkmark                      & \checkmark                  & \checkmark                        & \checkmark      \\
  Robustness                & $\times$    & \checkmark  & $\times$      & \checkmark                      & \checkmark                  & \checkmark                        & \checkmark      \\
  Compression               & $\times$    & $\times$    & $\times$      & \checkmark                      & \checkmark                  & \checkmark                        & \checkmark      \\
  Task awareness            & $\times$    & $\times$    & $\times$      & $\times$                        & $\times$                    & $\times$                          & \checkmark      \\
  Flexibility               & $\times$    & $\times$    & $\times$      & $\times$                        & $\times$                    & $\times$                          & \checkmark      \\
  Joint learning            & $\times$    & $\times$    & $\times$      & \checkmark                      & \checkmark                  & $\times$                          & \checkmark      \\ \bottomrule
  \end{tabular}}
\end{table*}
  
% --------------------------------------------------------------------------------------

\section{Hierarchical Multiscale Variational Autoencoder with Gaussian Mixture Model}
\subsection{Problem Statement}
\label{sec:event_triggered}
\textbf{Event-triggered time series:}
Here we take the traffic time series of IoT devices as an example to illustrate the temporal characteristics of such event-triggered traffic time series.

Let $\mathbf{x}^{\top} \in \mathbb{R}^{L}$ denote a traffic time series of length $L$ which is triggered by $K$ events, i.e, 
\begin{equation}
\label{eqn:events}
    \mathbf{x} = f(\mathbf{a}_1 \circ \mathbf{e}_1, \cdots, \mathbf{a}_K \circ \mathbf{e}_K),
\end{equation}
where $\mathbf{e}_i^{\top} \in \mathbb{R}^{L}$ is a indicator sequence, and its $t^{th}$ element, $e_{it} \in \{0,1\}$, signifies whether event $i$ is triggered at time $t$.
$\mathbf{a}_i^{\top} \in \mathbb{R}^{L}$ represents an intensity sequence, and its $t^{th}$ element, $a_{it} \in \mathbb{R}$, denotes the traffic generated by event $i$ at time $t$. 
The symbol $\circ$ denotes the Hadamard product.
The events discussed in this paper typically fall into two categories:
Machine-type communication (MTC) events, like transmitting sensing data and sending DNS requests, induce short-term and periodic dependencies. Meanwhile, Human Type Communication (HTC) events, such as requesting streaming videos or web pages via a smartphone, initiate long-term and bursty dependencies \cite{tahaei2020rise}.

Consider the traffic generated by a webcam. The traffic time series of the webcam is a superposition of traffic from $K$ events, represented as $\mathbf{x} = \sum_{i=1}^{K} \mathbf{a}_i \circ \mathbf{e}_i$, where interaction between the camera and controller, including routine queries and responses, are categorized as MTC events. These events, occurring frequently but consuming small traffic, can be considered background traffic, contributing to short-term dependencies in the time series.
In contrast, human interactions, such as viewing surveillance videos, are classified as HTC events. While less frequent than MTC events, HTC events consume significant traffic and manifest as bursts in the traffic time series.
The characteristics of MTC and HTC events are summarized in Table \ref{table:mtc_htc}.
     
\begin{table}[!tbp]
    \caption{Characteristics of MTC and HTC events}
    \label{table:mtc_htc}
    \centering
    \begin{tabular}{@{}ccl@{}}
        \toprule
        Event Type & Volume & \multicolumn{1}{c}{Pattern}                                                                             \\ \midrule
        MTC        & Small  & \begin{tabular}[c]{@{}l@{}}Almost periodic\\ Short transmission period\end{tabular}                     \\
        HTC        & Large  & \begin{tabular}[c]{@{}l@{}}Bursty and unpredictable\\ Generally long transmission interval\end{tabular} \\ \bottomrule
    \end{tabular}
\end{table}

Event-triggered time series are primarily characterized by \textit{heterogeneity}, and \textit{dependency across multiple time scales}. HTC events, originating from various applications and connecting to different servers, along with the unpredictability of human behavior, lead to significant heterogeneity even within the normal data. Furthermore, traffic from a single MTC event is notably homogeneous and typically periodic \cite{tahaei2020rise}. Due to the superposition of multiple events, the seasonality stemming from MTC events is observable across different time scales, resulting in an intricate dependency.

\textbf{Problem Statement}: Given a collection of event-triggered time series denoted by $\mathbf{X} = \left[ \mathbf{x}_1, \mathbf{x}_2, \cdots, \mathbf{x}_N\right]$, which may exhibit high dynamics, heterogeneity, and variation in time granularity. Our objective is twofold: (1) to obtain a robust low-dimensional representation vector for each time series in $\mathbf{X}$ for similarity measurement; (2) to detect anomalous time series $\mathbf{x}_i$ or to cluster the all time series based on the learned representations and similarity measurements.

We will describe the architecture of the basic ensemble component in Section \ref{sec:comp}. Then, we introduce the W and D compression networks in Sections \ref{sec:w_comp} and \ref{sec:d_comp}, which are the main differences between the W and D branches. Other technical details are described in Section \ref{sec:others}.

\begin{figure*}
\centering
    \includegraphics[width=2\columnwidth]{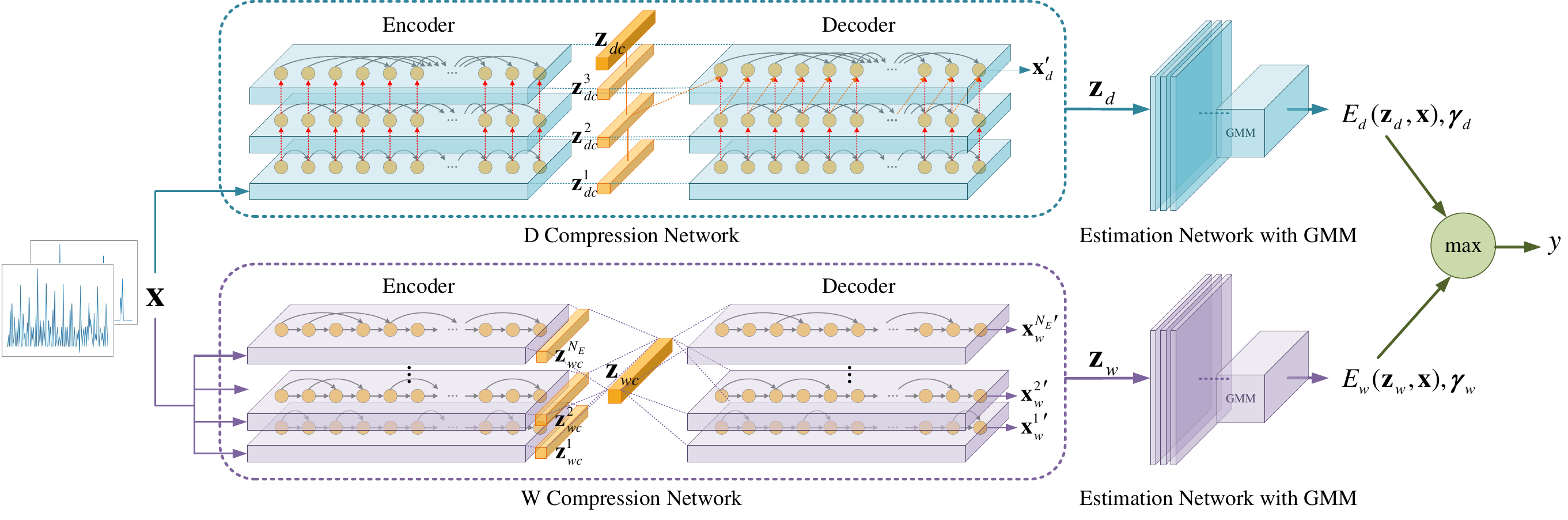}
    \caption{The architecture of ET-Net}
\label{fig:framework}
\end{figure*}

\subsection{General Framework}
\label{sec:comp}

The basic ensemble component consists of two modules: a compression network and a distribution estimator. The compression network maps the event-triggered time series into a low-dimensional latent space $\mathcal{Z}$, while the distribution estimator models the probability distribution of the latent space using GMM. These two modules work in a coordinated manner to jointly capture the temporal dynamics of the event-triggered time series and generate vector embeddings optimized for GMM. The objective function used to learn this model is presented below.

\begin{equation}
\label{general_eqn:loss}
    L = \left\| {\mathbf{X} - g(\mathbf{X})} \right\|_2 ^2  + \lambda E(\mathbf{Z}, \mathbf{X}),
\end{equation}
where $g(\cdot)$ is the autoencoder. $\mathbf{Z}$ is the representation of $\mathbf{X}$ in the latent space. $\left\|  \cdot  \right\|_2 ^2$ denotes the squared L2-loss.
$E(\cdot)$ is the negative log-likelihood of the estimated GMM, a.k.a an energy function, which models latent space distribution.
$\lambda$ is a weighting parameter that governs the tradeoff between two individual objective functions.
The above formula is similar to that of various existing works, such as VAE \cite{kingma2013auto}, DAGMM \cite{zong2018deep}, DTCR \cite{ma2019learning}, and SOM-VAE \cite{fortuin2018som}.

\textcolor{black}{
As previously mentioned, the heterogeneity and complex temporal dependencies in event-triggered time series pose challenges to representation learning.
In this section, we propose two special types of sequence auto-encoder architectures that are partly similar to seq2seq \cite{sutskever2014sequence} and are particularly suitable for learning the robust representation of event-triggered time series. We first propose the W compression network, which is inspired by multi-task learning, to tackle the heterogeneity. This network treats recurrent networks with varying time scales as independent tasks, jointly optimized for robust representation of heterogeneous sequences.
To address the complex temporal dependencies, we developed a D compression network, inspired by multi-resolution learning. This network incrementally learns temporal representations, transitioning from fine-grained to coarse-grained.
To avoid the curse of dimensionality, we equip each compression network with an estimation network featuring a learnable GMM, rather than directly concatenating the two representations. This approach constrains the sample within a GMM.
We further ensemble these two anomaly detection models using the bagging method, aiming to enhance anomaly detection performance.
}

\subsection{W compression network}
\label{sec:w_comp}

The motivation behind designing a W compression network is to capture the intricate temporal dependencies among event-triggered time series in a multi-task learning fashion, which has been demonstrated to yield robust representations \cite{long2017learning,mao2020multitask}. This is particularly suitable for event-triggered time series, which can be highly heterogeneous. To achieve this, each encoder-decoder pair in the W compression network randomly models a different type of temporal dependency with a distinct time span using a Stochastic Recurrent Neural Network (SRNN) \cite{kieu2019outlier}. The output of the W compression network includes a compressed latent space representation $\mathbf{z}_{wc}$ and an extended latent space representation $\mathbf{z}_{w}$. The architecture of the W compression network is illustrated in Figure \ref{fig:framework}.

Reconstructed sequence output by $i^{th}$ decoder $\mathbf{x}^{i\prime}_{w}$ is computed as
$\mathbf{z}^{i}_{wc} = g^{i}_{we} (\mathbf{x})$ and $\mathbf{x}^{i\prime}_{w}  = g^{i}_{wd} (\mathbf{z}_{wc})$,
where $g_{we}^{i}(\cdot)$ and $g_{wd}^{i}(\cdot)$ are the $i^{th}$ encoder and decoder of W branch, $\mathbf{z}^{i}_{wc}$ is the final state of $i^{th}$ encoder. $\mathbf{z}_{wc} = {\mathbf{W}}_{w} [\mathbf{z}^{1}_{wc}, \cdots, \mathbf{z}^{N_E}_{wc}]^T + \mathbf{b}_{w}$,
where $N_E$ is the number of encoders/decoders, $\mathbf{W}_{w}$ and
$\mathbf{b}_{w}$ denote a trainable weight matrix and a bias vector,
respectively.
The extended latent space representation is then given by
$\mathbf{z}_{w} = [\mathbf{z}_{wc} , d_{rel}(\mathbf{x}, \mathbf{x}^{p\prime}_{w}), d_{cos} (\mathbf{x},\mathbf{x}^{q\prime}_{w}]$, where $d_{rel}(\cdot)$ and $d_{cos}(\cdot)$ are the reconstruction error, denoting the relative distance and cosine similarity, respectively, $p$ and $q$ are the indexes of auto-encoder branch with the minimum reconstruction relative distance and cosine distance, respectively.

The \textit{recurrent function} used to update the hidden state of the recurrent cell in the $i$-th layer of is as follows,
\begin{equation}
\label{eqn:srnn}
    \begin{array}{l}
        \mathbf{h}^i (t)  = \frac{ w_1^i(t) \cdot f_{rnn} (\mathbf{h}^i (t-1) ,x(t) ) + w_2^i(t)  \cdot f'(\mathbf{h}^i(t-s^i) ,x(t) )}{w_1^i(t) + w_2^i(t)} \\
        s.t.\quad w_1^i(t) ,w_2^i(t)  \in \{ 0,1\} , w_1^i(t)  + w_2^i(t)  \ne 0 \\
    \end{array}
\end{equation}
where $w_1^i(t)$ and $w_2^i(t)$ are randomly initialized weights.
$f_{rnn} (\cdot)$ denotes a non-linear function including Long-Short Term Memory (LSTM)
\cite{hochreiter1997long} or Gated Recurrent Unit (GRU) \cite{chung2014empirical}.
$f'(\cdot)$ denotes a linear operation.
$s^i$ is a parameter that controls the memory ability of SRNN.
When $s^i$ is small, SRNN tends to learn short-term dependencies.
Otherwise, it will learn long-term dependencies.
We set the parameter $s^i$ of each encoder/decoder to a different value but no more than three, so that it tends to learn short-term dependencies in time series.

When LSTM is set as the recurrent cell, $f_{rnn}(\cdot)$ in (\ref{eqn:srnn}) can be expanded as $f_{lstm}(\cdot)$,
\begin{equation}
\label{eqn2}
    \begin{array}{l}
        f_{lstm} (\mathbf{h}(t - 1),\mathbf{c}(t - 1),x(t)) = \mathbf{o}(t) \circ  \mathbf{c}(t) \\
        \mathbf{o}(t) = \sigma (\mathbf{W}_o  \cdot [\mathbf{
        h}(t - 1) , x(t)] + \mathbf{b}_o ) \\
        \mathbf{c}(t) = \mathbf{f}(t) \circ \mathbf{c}(t - 1) + \mathbf{i}(t) \circ \mathbf{\tilde c}(t) \\
        \mathbf{f}(t) = \sigma (\mathbf{W}_f  \cdot [\mathbf{h}(t - 1) , x(t)] + \mathbf{b}_f ) \\ 
        \mathbf{i}(t) = \sigma (\mathbf{W}_i  \cdot [\mathbf{h}(t - 1) , x(t)] + \mathbf{b}_i ) \\
        \mathbf{\tilde c}(t) = \tanh (\mathbf{W}_c  \cdot [\mathbf{h}(t - 1) , x(t)] + \mathbf{b}_c )
    \end{array}
\end{equation}  
where $\mathbf{i}$, $\mathbf{o}$ and $\mathbf{f}$ are input gate, output gate and forget gate respectively. $\mathbf{c}$ is the memory.
$\mathbf{W}_o$, $\mathbf{W}_f$, $\mathbf{W}_i$, $\mathbf{W}_c$, $\mathbf{b}_o$, $\mathbf{b}_f$, $\mathbf{b}_i$ and $\mathbf{b}_c$ are the parameters to be learned. The $f_{rnn}(\cdot)$ is defined as $f_{gru}(\cdot)$ when GRU is set as the recurrent cell.
\begin{equation}
\label{eqn4}
    \begin{array}{l}
        f_{gru} (\mathbf{h}(t - 1),x(t)) = (\mathbf{1} - \mathbf{u}(t)) \circ \mathbf{h}(t - 1) + \mathbf{u}(t) \circ \mathbf{ \tilde h}(t) \\
        \mathbf{u}(t) = \sigma (\mathbf{W}_u  \cdot [\mathbf{h}(t - 1) , x(t)] + \mathbf{b}_u) \\
        \mathbf{ \tilde h}(t) = \tanh (\mathbf{W}_h  \cdot [(\mathbf{r}(t) \circ \mathbf{h}(t - 1)) ,  x(t)] + \mathbf{b}_h) \\
        \mathbf{r}(t) = \sigma (\mathbf{W}_r  \cdot [\mathbf{h}(t - 1) , x(t)] + \mathbf{b}_r)
    \end{array}
\end{equation} 
where ${{{\mathbf{u}}}}$ and ${{{\mathbf{r}}}}$ are update gate and reset gate respectively,
${\mathbf{W}}_u$, ${\mathbf{W}}_h$, ${\mathbf{W}}_r$, ${{{\mathbf{b}}}}_u$, ${{{\mathbf{b}}}}_h$ and ${{{\mathbf{b}}}}_r$ are parameters to be learned.

\subsection{D compression Network}
\label{sec:d_comp}

The purpose of the D compression network is to use multi-resolution learning to capture the complex long-term dependencies (introduced by HTC) and short-term dependencies (introduced by MTC) in event-triggered time series. To achieve this, multiple layers of dilated RNNs \cite{chang2017dilated} are stacked sequentially to obtain deep representations of time series. The D compression network outputs a compressed vector $\mathbf{z}_{dc}$ and an extended latent representation $\mathbf{z}_d$ that can be used for further processing. Figure \ref{fig:framework} illustrates the architecture of the D compression network.

The D compression network learns representations of the time series at different levels of granularity, from fine-grained to coarse-grained. The final state of each encoder layer is denoted as $\mathbf{z}^{i}_{dc}$, and the overall compressed vector is computed as $\mathbf{z}_{dc} = \mathbf{W}_{d} \cdot [\mathbf{z}^{1}_{dc}, \cdots ,\mathbf{z}^{N_L}_{dc}]^T + \mathbf{b}_{d}$, where $\mathbf{W}_{d}$ and $\mathbf{b}_{d}$ are trainable weight and bias parameters, respectively. ${N_L}$ is the number of resolution levels. The resulting $\mathbf{z}_{dc}$ is then used as the input for the decoder, which generates the reconstructed time series $\mathbf{x}^{\prime}_{d}$. The extended representation $\mathbf{z}_{d}$ is then formed by concatenating $\mathbf{z}_{dc}$ with the relative distance $d_{rel} (\mathbf{x}, \mathbf{x}^{\prime}_{d})$ and the cosine similarity $d_{cos} (\mathbf{x}, \mathbf{x}^{\prime}_{d})$.

The hidden state of the recurrent cell in the $i$-th layer of dilated RNN is updated as
\begin{equation}
\label{eqn:drnn}
    \begin{array}{l}
        \mathbf{h}^i (t) = f(\mathbf{h}^{i-1} (t) ,\mathbf{h}^i (t-d^i) ) \\
        \mathbf{h}^0 (t) = x(t)
    \end{array}
\end{equation}
where $d^i$ denotes the dilation size in $i^{th}$ layer. The hidden state of layer $i$ at time $t$ only depends on the state at $t-d^i$ and the fine-grained state at layer $i-1$. In practice, we set 3 as the dilations in first layer, then an exponential growth strategy is used to set the dilation in the subsequent layer, that is, $d^i = 3^{i}$.

\subsection{Estimation Network, Loss Function and Task-aware Output}
\label{sec:others}

\subsubsection{Estimation Network}

After obtaining the extended latent space representation $\mathbf{z}_{w}$ or $\mathbf{z}_{d}$, it is input to a GMM estimator for density estimation, as illustrated in the following equation. To simplify the notation, we omit the subscripts and use $\mathbf{z}$ to represent the extended latent space representation generated by either the W or D compression network.

\begin{equation}
\label{eqn:2}
    \begin{array}{l}
        \boldsymbol{\gamma} = g_m\left( \mathbf{z} \right)\quad
        \varphi _k  = \sum\nolimits_{i = 1}^M {\frac{{\boldsymbol{\gamma} _{ik} }}{M}}  \quad
        \mu _k ,\Sigma _k  = \eta \left( {\{ [\mathbf{z}_i,\boldsymbol{\gamma}_i ]\} _{i = 1}^M } \right),
    \end{array}
\end{equation}
where $K$ is the number of mixture components in GMM and $M$ is the number of samples in mixture component $k$. $g_m(\cdot)$ is a membership estimator, and $\boldsymbol{\gamma}$ is a $K$-dimensional vector representing the probability that sample $\mathbf{z}$ belonging to the $k^{th}$ mixture component.
$\varphi _k$, $\mu _k$ and $\Sigma _k$ are mixture probability, mean and covariance for $k^{th}$ mixture component, respectively.
$\eta(\cdot)$ denotes a function for computing the mean and covariance.
In practice, we use the iterative EM algorithm to update $\mu _k$ and $\Sigma _k$ based on $\mathbf{z}$, instead of computing them directly using $\boldsymbol{\gamma}$ and $\boldsymbol{\varphi}$ like DAGMM.

Once we obtain the parameters of the GMM model, the sample energy function (c.f. (\ref{general_eqn:loss})) can be calculated as follows,
\begin{equation}
\label{eqn:energy}
    \begin{array}{l}
        E (\mathbf{z} , \mathbf{x}) =  - \log \left( {\sum\limits_{k = 1}^K {\varphi _k \theta \left( {\mathbf{z}|(\mu _k ,\Sigma _k )} \right)} } \right) \\
        \theta \left( {\mathbf{z}, (\mu _k ,\Sigma _k )} \right) = \frac{1}{{\sqrt {\left| {2\pi \Sigma _k } \right|} }}\exp \left( { - \frac{{(\mathbf{z} - \mu _k )^2 }}{{2\Sigma _k }}} \right)
    \end{array}
\end{equation}
where the $ \theta \left( {\mathbf{z}, (\mu _k ,\Sigma _k )} \right)$ is the density function of Gaussian distribution ${\cal N}(\mu _k ,\Sigma _k )$.

\subsubsection{Loss Function and Task-aware Output}
During training, ET-Net is trained end-to-end using either normal or unlabeled data. It's assumed that the unlabeled data contains few anomalies, an assumption typically holds due to the rarity of anomalies.
The loss function for the W branch and the D branch is given below, where $N$ is the number of training samples.

\begin{equation}
    \begin{array}{rl}
    L_w & = \frac{1}{N N_E} \sum\limits_{i = 1}^N {\sum\limits_{j = 1}^{N_E} { \left\| {\mathbf{x}_i - \mathbf{x}^{j\prime}_{wi} } \right\|_2 ^2 }} + \frac{\lambda}{N} \sum\limits_{i = 1}^N {E_w(\mathbf{z}_{wi} , \mathbf{x}_{i})} \\
    
    L_d & = \frac{1}{N} \sum\limits_{i = 1}^N {\left\| { \mathbf{x}_i -  \mathbf{x}^{\prime}_{di}} \right\|_2 ^2 }  + \frac{\lambda}{N} \sum\limits_{i = 1}^N {E_d(\mathbf{z}_{di} , \mathbf{x}_i)}.
    \end{array}
\end{equation}
where $E_{w}(\cdot)$ and $E_{d} (\cdot)$ denote the energy functions of W and D branches, respectively.

During testing, the output of ET-Net $y$ corresponds to the specific machine learning task we aim to carry out. For anomaly detection, $y$ represents the anomaly score of the sample $\mathbf{x}$, and is calculated as $y = \max (E_{w} (\mathbf{z}_w, \mathbf{x}),E_{d} (\mathbf{z}_d, \mathbf{x}))$,  In doing so, the network outputs the highest anomaly score to ensure high recall.
For clustering or classification tasks, $y$ represents the predicted label and is defined as $y = \arg \max _i ( \max([\boldsymbol{\gamma}_w,\boldsymbol{\gamma}_d]^{\top}) )$, where ${{\boldsymbol{\gamma}_w}}$ and ${{\boldsymbol{\gamma}_d}}$ represent probabilistic GMM membership predicted by W and D branches respectively. Here we take the maximum of the two predicted probabilities since the resulting output by the sharper softmax distribution is preferred \cite{hendrycks2016baseline}.

\section{Experiments}

\textcolor{black}{Recall that the motivation of this paper is to detect anomalous or malicious behavior of devices or software via their traffic time series. Besides focusing on anomaly detection, it also explores the underlying fundamental problem of similarity learning. Consequently, the experiment considers two security-related machine learning tasks, i.e., anomaly detection and clustering. The applications of these two tasks in security management include intrusion detection and device identification. Furthermore, the clustering results also aim to validate the adaptability of similarity learned by ET-Net on tasks other than anomaly detection.}
We carry out the study to answer the following questions regarding the proposed approach. 
\begin{itemize}
    \item \textbf{Effectiveness}: Whether ET-Net outperforms the existing state-of-the-art anomaly detection and clustering methods?
    \item \textbf{Robustness}: Whether the trained model capable of being robust to noise, time granularity variations and potential traffic disturbance, which often occur during practical deployment?
    \item \textbf{Visualization}: Can we visualize and interpret the similarity metric learned by ET-Net?
\end{itemize}

\subsection{Datasets and Experimental Design}
\subsubsection{Datasets} 
We first conduct anomaly detection and clustering experiments with synthetic datasets to visualize the latent space learned by ET-Net, which provides an intuitive interpretation of the model results.

We then conduct anomaly detection on several public, real-world traffic datasets. The UNSW-IoT \footnote{\url{https://iotanalytics.unsw.edu.au/iottraces.html}} \cite{sivanathan2018classifying},
cell traffic \footnote{\url{https://dandelion.eu/datagems/SpazioDati/telecom-sms-call-internet-mi}},
IoT23\footnote{\url{https://www.stratosphereips.org/datasets-iot23}},
and PowerCons dataset from the UCR time series classification archives\footnote{\url{https://www.cs.ucr.edu/~eamonn/time_series_data_2018}} are four event-triggered time series datasets.

\textcolor{black}{The UNSW-IoT and IoT23 datasets originate from traffic data of IoT devices. These devices generate traffic only when they communicate, and remain in a dormant state at other times. The communication events include HTC and MTC. During the collection process, packets from all devices are captured and recorded, with each device being identified by a unique label. We divide the traffic time series into non-overlapping windows, each covering a 120-minute interval. Within each window, each data point represents the number of packets collected in one minute. For the UNSW-IoT dataset, the targeted anomalous events for detection are communications initiated directly by humans on non-IoT devices. These events pose potential security risks to private data in the sensing network \cite{ortiz2019devicemien,sivanathan2017characterizing}. In the IoT23 dataset, our objective is to detect malicious attack events within the sensing network.}

\textcolor{black}{
The cell traffic dataset is derived from traffic data on the cellular base station, which generates traffic only when users within the coverage area initiate communication activities. If no users are utilizing the mobile network during a certain period, the data will be zero. Time series data from a selected cell are considered non-anomalous. We create anomalies by injecting traffic time series from another cell. Thus, anomalous events are defined as communications that deviate from a cell's historical patterns, possibly signifying unusual public events.}

For the UNSW-IoT, IoT23, and cell traffic datasets, we divided them into training and test datasets with a 40-60 split. The proportion of normal and abnormal samples in these datasets are 0.012, 0.093, and 0.077, respectively.

\textcolor{black}{The PowerCons dataset originates from household electricity usage, recording data solely during human electricity use. In this dataset, data from one season is considered normal, while data from other seasons are treated as anomalies \cite{benkabou2018unsupervised}. Anomalous events in this dataset are defined as consumption behaviors that deviate from a season's typical pattern, potentially indicating illegal activities such as unauthorized bitcoin mining.}

\textcolor{black}{
In addition, we selected three non-event-triggered datasets from the UCR time series classification archive, MedicalImages, SmoothSubspace, and MoteStrain, as these datasets have similar heterogeneous properties as event-triggered time series. Additionally, the results on these datasets emphasize the adaptability of ET-Net to a wide range of applications. }
We followed the anomaly injection experimental setup detailed in \cite{benkabou2018unsupervised} for these datasets, as well as for the PowerCons dataset. This approach ensures the abnormal-to-normal sample ratio does not exceed 0.1.
AUC (Area under the receiver operating curve) is employed to assess the anomaly detection performance.

Likewise, the clustering performance of the proposed framework is elucidated via experiments on three real-world datasets, including UNSW-IoT, IoT23 and cell traffic. NMI (Normalized Mutual Information) is employed to assess the clustering performance.

\subsubsection{Baselines}
For anomaly detection, we compare ET-Net against the following state-of-the-art unsupervised methods, including One-Class SVM (OCSVM),
Local Outlier Factor (LoF), Isolation Forest (IF), Dynamic Time Warping (DTW), KitNET \cite{mirsky2018kitsune}, GRU-AE \cite{malhotra2016lstm}, Shared-SRNN \cite{kieu2019outlier}, DAGMM \cite{zong2018deep}, and USAD \cite{audibert2020usad}.

The baseline algorithms for clustering include K-means, GMM, K-means+DTW, K-means+EDR, K-shape \cite{paparrizos2015k}, DEC \cite{xie2016unsupervised}, IDEC \cite{guo2017improved}, SPIRAL \cite{lei2019similarity}, DTC \cite{madiraju2018deep}, and Autowarp \cite{abid2018learning}.

For a fair comparison, all baseline methods use the parameter settings recommended by the authors.

\subsection{Latent Space Visualization on Synthetic Datasets}
In this section, we visualize the latent space learned by ET-Net on anomaly detection and clustering tasks to elucidate the underlying mechanism of the ET-Net, and provide an intuitive explanation for the model outcomes. Finally, we examine the robustness of latent space representation against data granularity variations and different types of noise. 

\textcolor{black}{To assess visualization results quantitatively, we utilize the average Silhouette Coefficients (SC) in clustering tasks. An SC closer to 1 indicates better performance, whereas an SC approaching -1 signifies poorer clustering. For the unsupervised anomaly detection task, ET-Net uses a GMM to group normal samples in the latent space and identifies samples that deviate from the normal cluster as anomalies. Ideally, the samples within the normal cluster should be highly cohesive and deviate from the scattered anomalous samples, so we use the Silhouette Coefficient Ratio (SCR) for normal and abnormal clusters to evaluate the visualization results. An SCR greater than 1, with higher values, indicates better performance, while an SCR close to 1 signifies ineffective separation between normal and abnormal clusters.}

\subsubsection{Anomaly detection}
\begin{figure}[!tbp]
    \centering
    \subfloat[Sine wave]{\includegraphics[width=0.31\columnwidth]{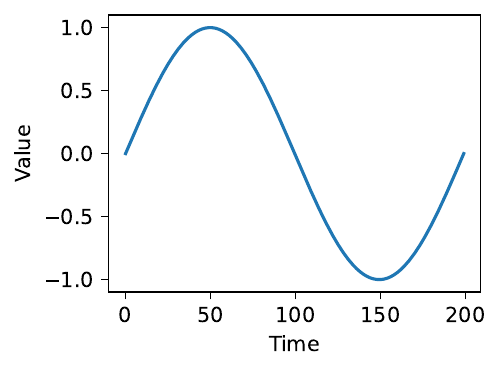}}
    \subfloat[Square wave]{\includegraphics[width=0.31\columnwidth]{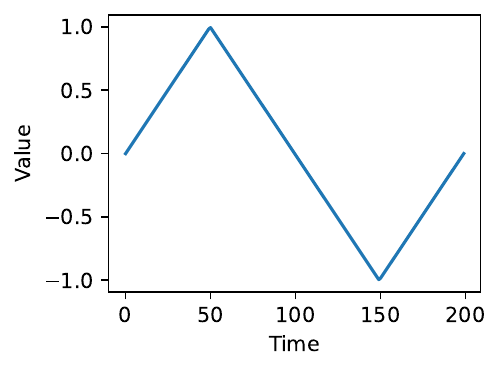}}
    \subfloat[Triangle wave]{\includegraphics[width=0.31\columnwidth]{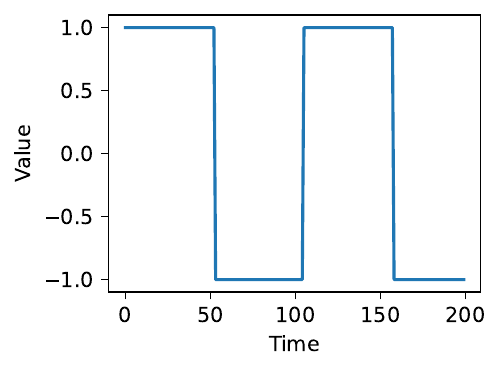}}
    \caption{\textcolor{black}{Normal samples from the synthetic anomaly detection dataset that exhibit temporal heterogeneity.}}
    \label{fig:exp_anomaly}
\end{figure}

We first assess the performance of the proposed framework via conducting machine learning tasks on a synthetic dataset, as shown in Figure \ref{fig:exp_anomaly}. This dataset consists of three non-anomalous time series samples, i.e., a sine wave, a square wave, and a triangle wave. We also create a total of five hundred copies for each time series and use them to train the proposed deep learning model.

\begin{figure*}
    \centering
    \subfloat[Type-1 Anomaly, SCR of 2D visualization is 27.36]{%
        \begin{minipage}{0.48\textwidth}
            \centering
            \includegraphics[width=0.2\linewidth]{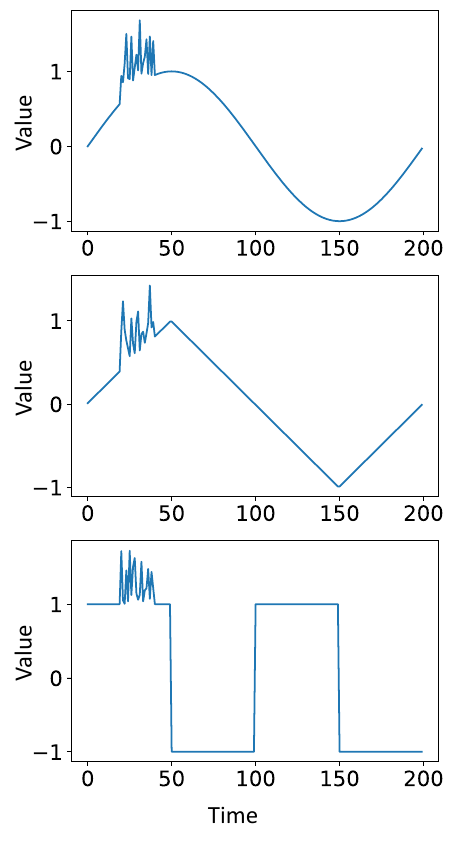}\hfil
            \includegraphics[width=0.39\linewidth]{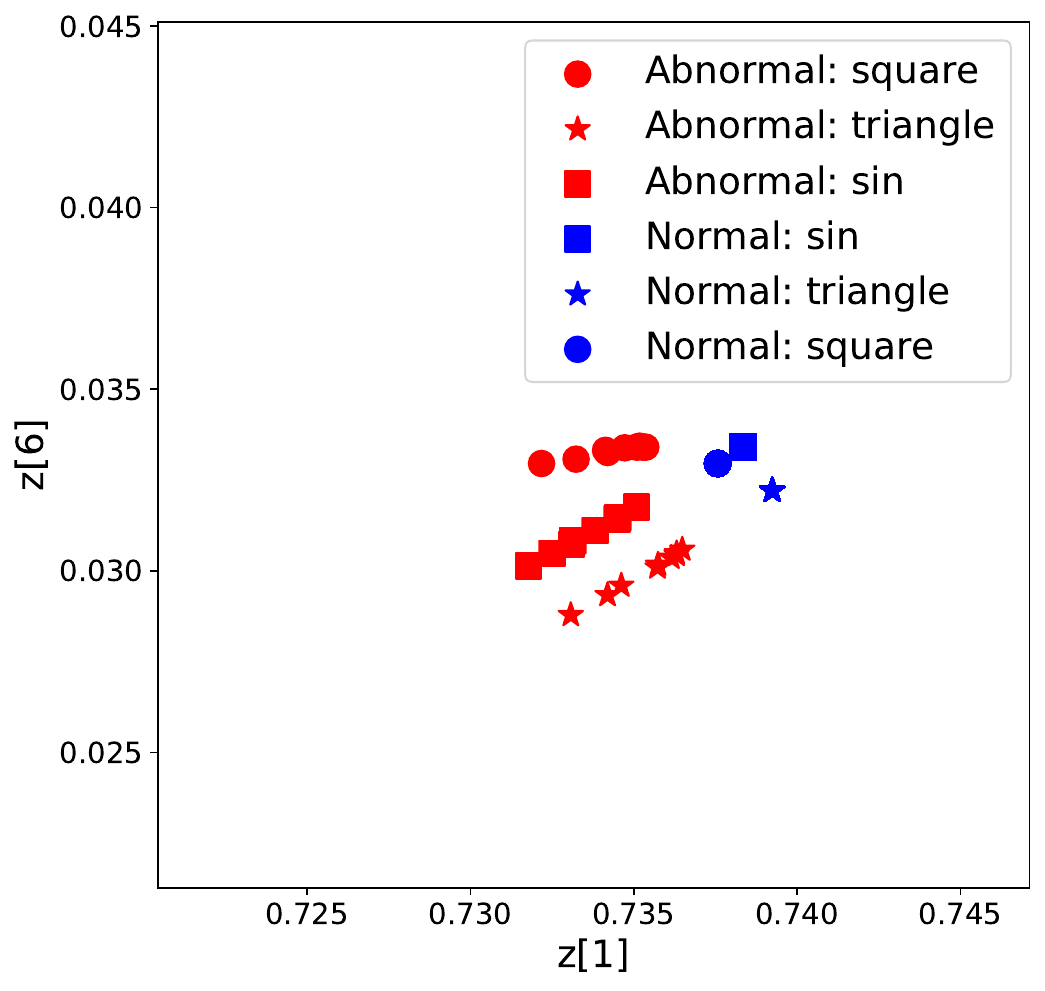}\hfil
            \includegraphics[width=0.39\linewidth]{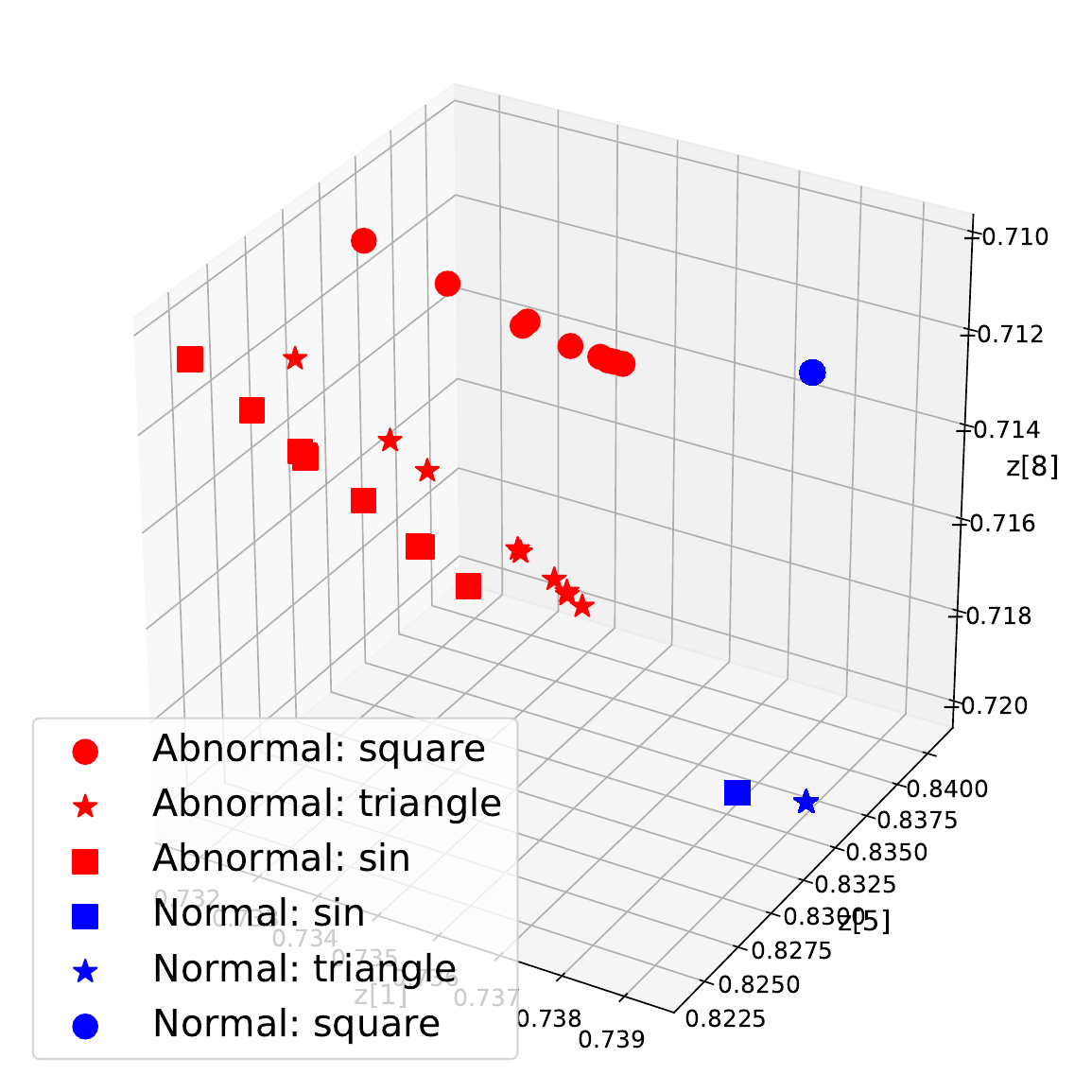}
        \end{minipage}%
    }\hfil
    \subfloat[Type-2 Anomaly, SCR of 2D visualization is 136.71]{%
        \begin{minipage}{0.48\textwidth}
            \centering
            \includegraphics[width=0.2\linewidth]{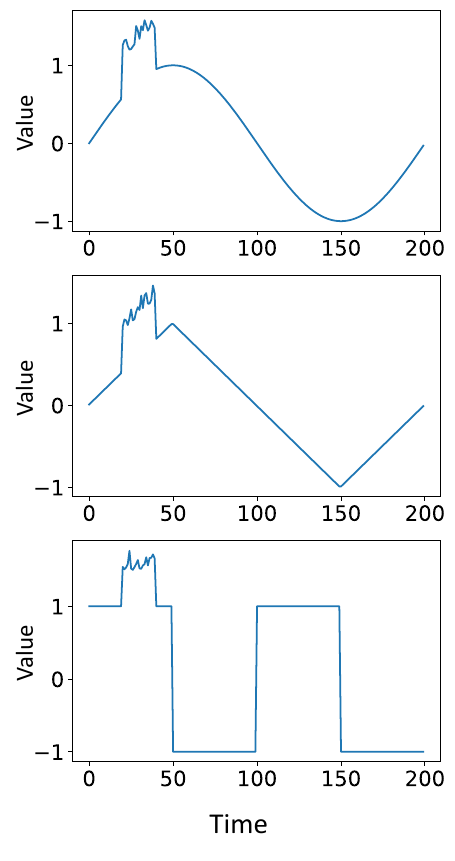}\hfil
            \includegraphics[width=0.39\linewidth]{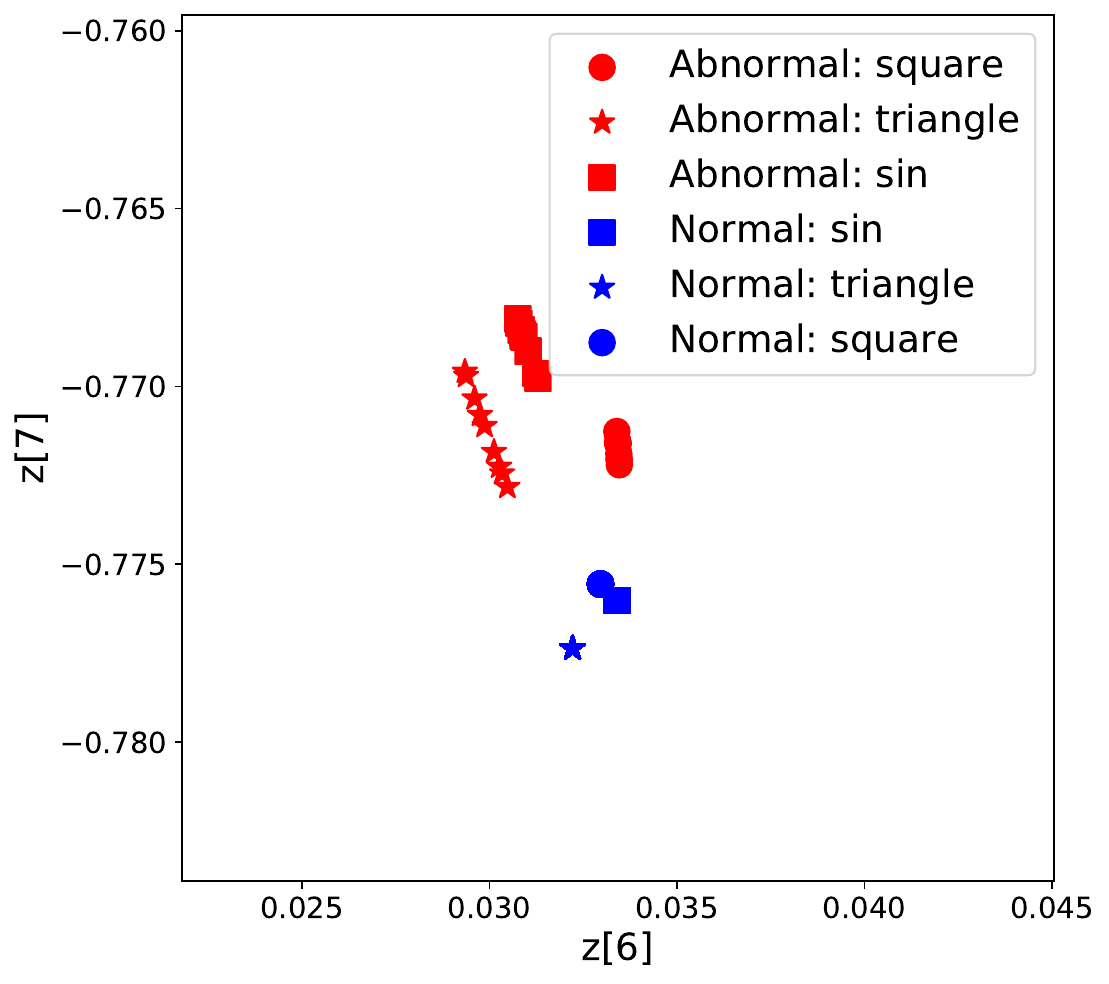}\hfil
            \includegraphics[width=0.39\linewidth]{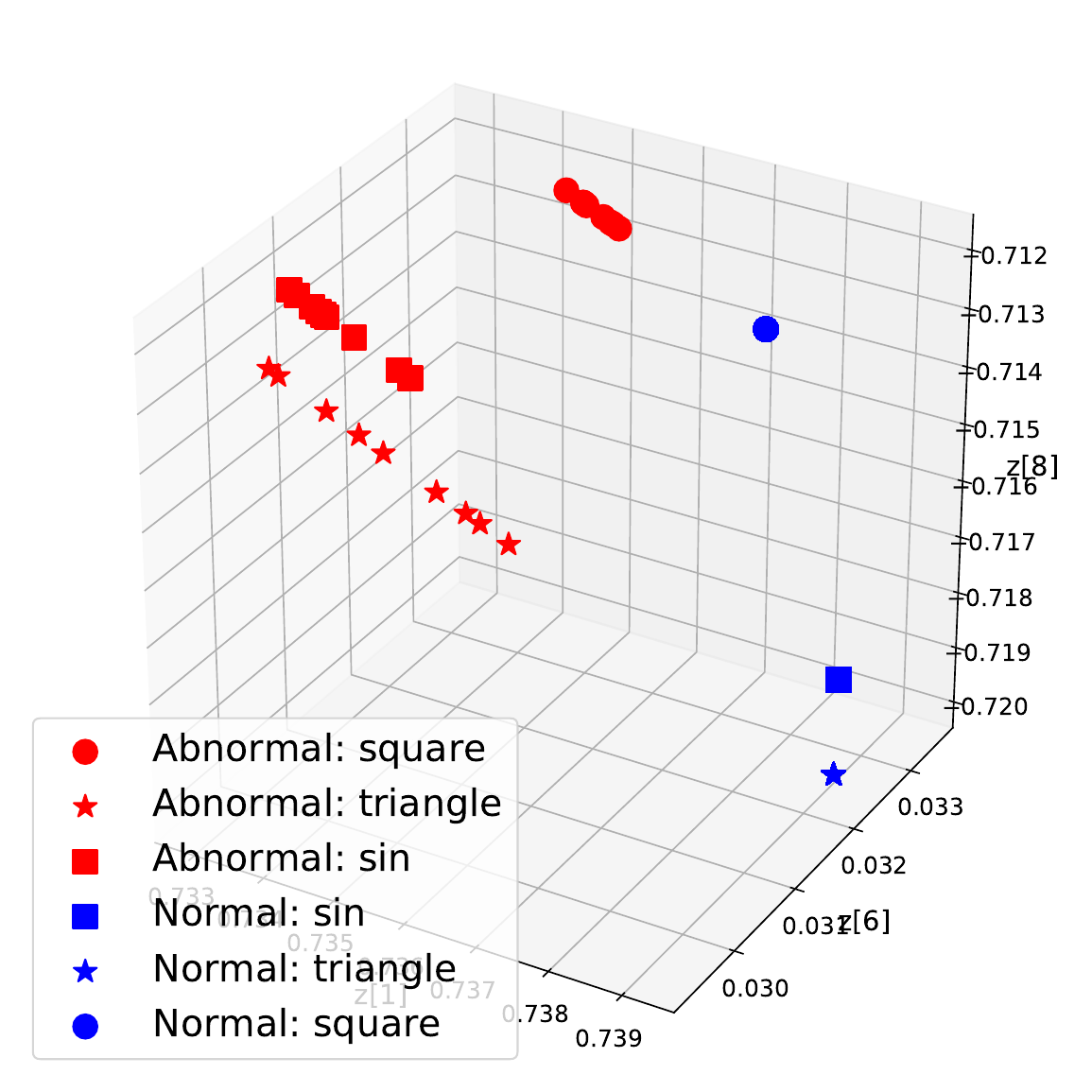}
        \end{minipage}%
    } \\ % \\ 用于换行

    \subfloat[Type-3 Anomaly, SCR of 2D visualization is 15.00]{%
        \begin{minipage}{0.48\textwidth}
            \centering
            \includegraphics[width=0.2\linewidth]{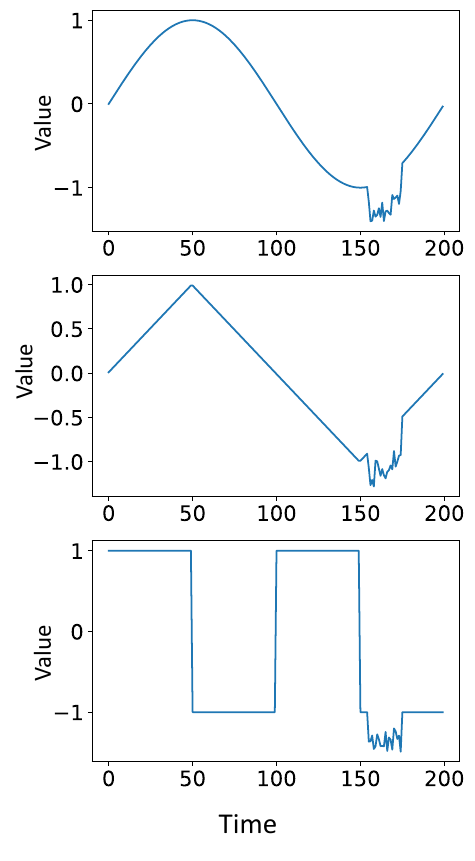}\hfil
            \includegraphics[width=0.39\linewidth]{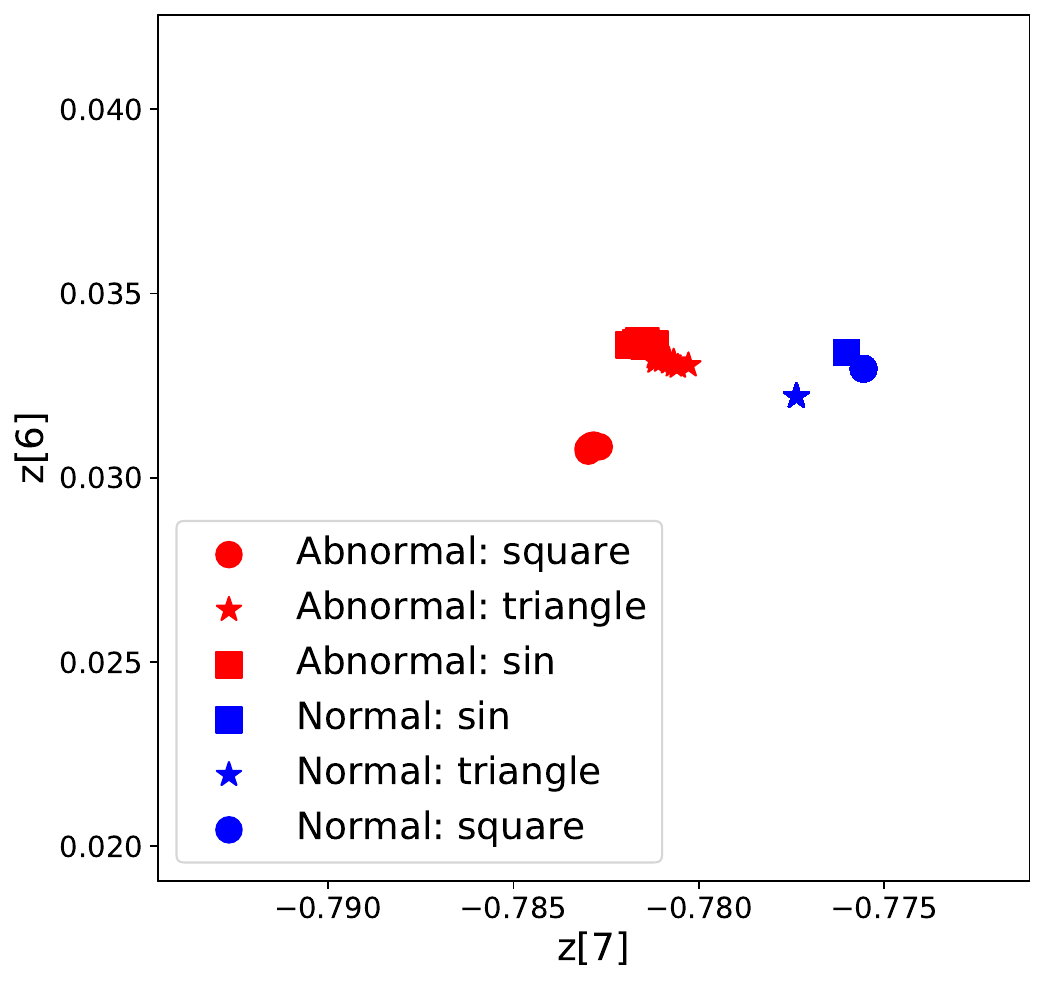}\hfil
            \includegraphics[width=0.39\linewidth]{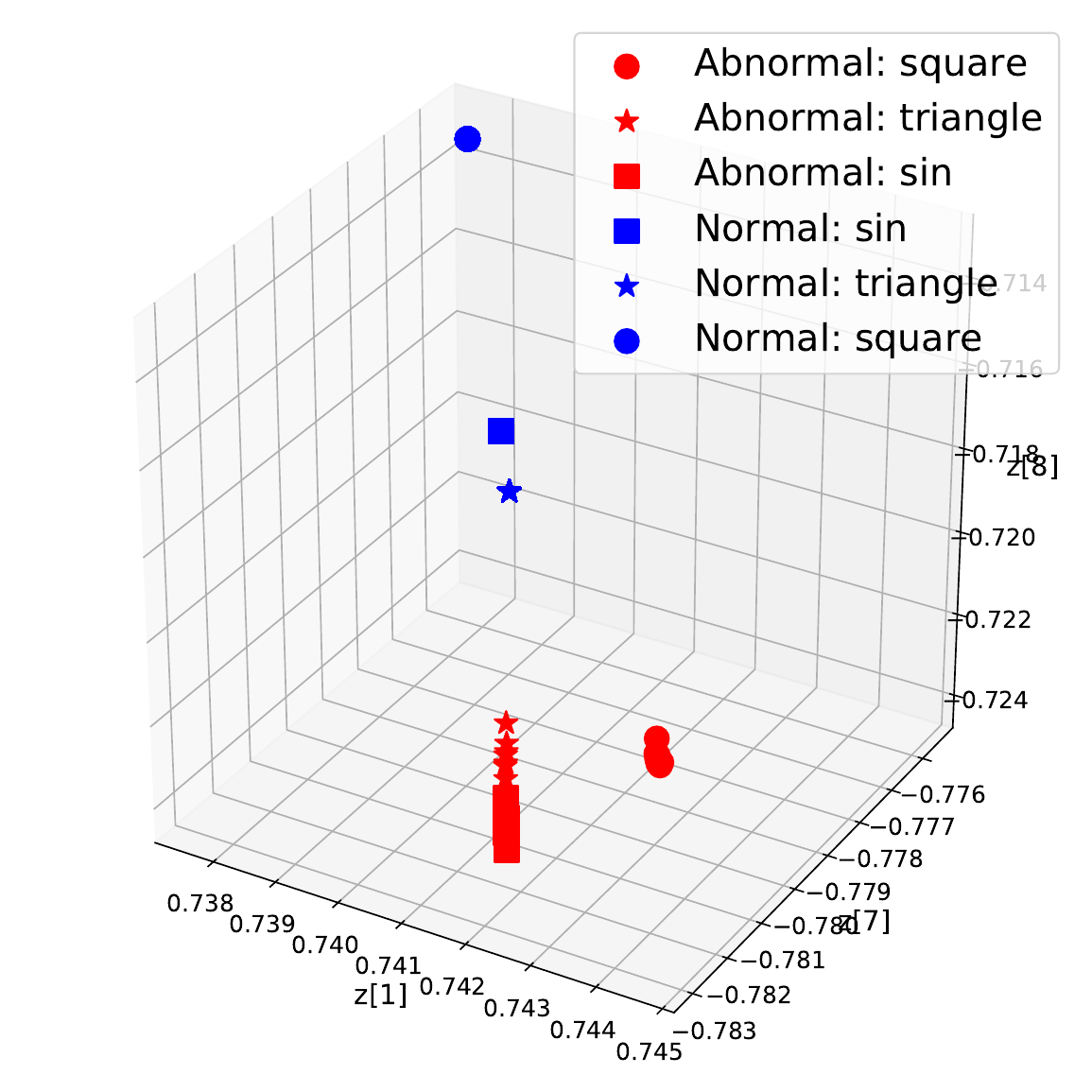}
        \end{minipage}%
    }\hfil
    \subfloat[Type-4 Anomaly, SCR of 2D visualization is 50.55]{%
        \begin{minipage}{0.48\textwidth}
            \centering
            \includegraphics[width=0.2\linewidth]{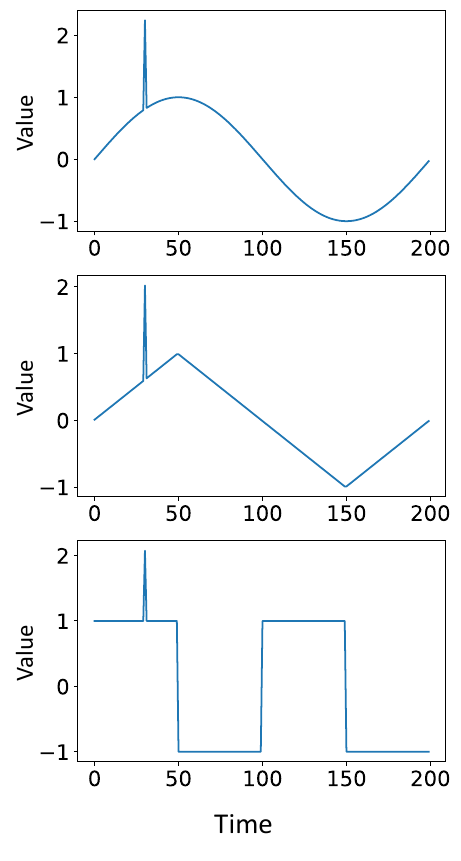}\hfil
            \includegraphics[width=0.39\linewidth]{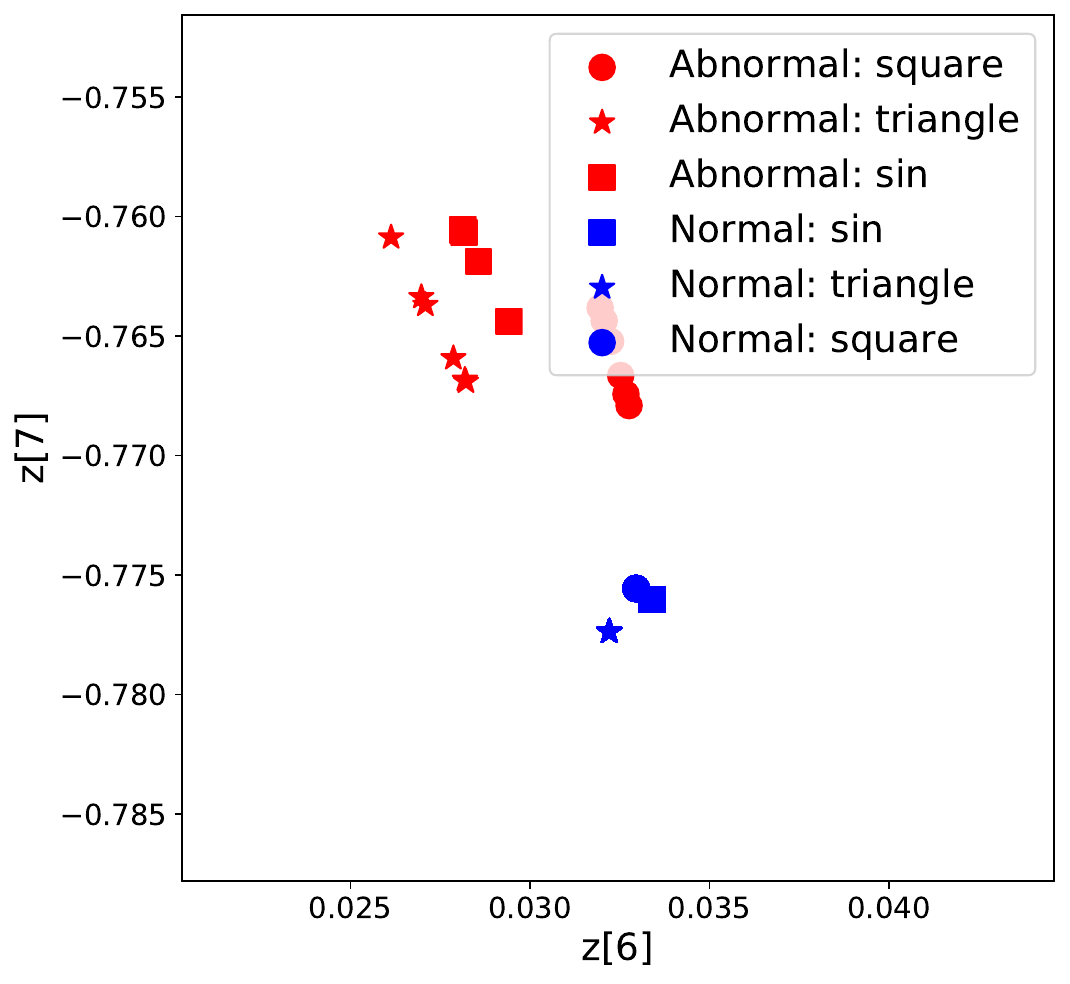}\hfil
            \includegraphics[width=0.39\linewidth]{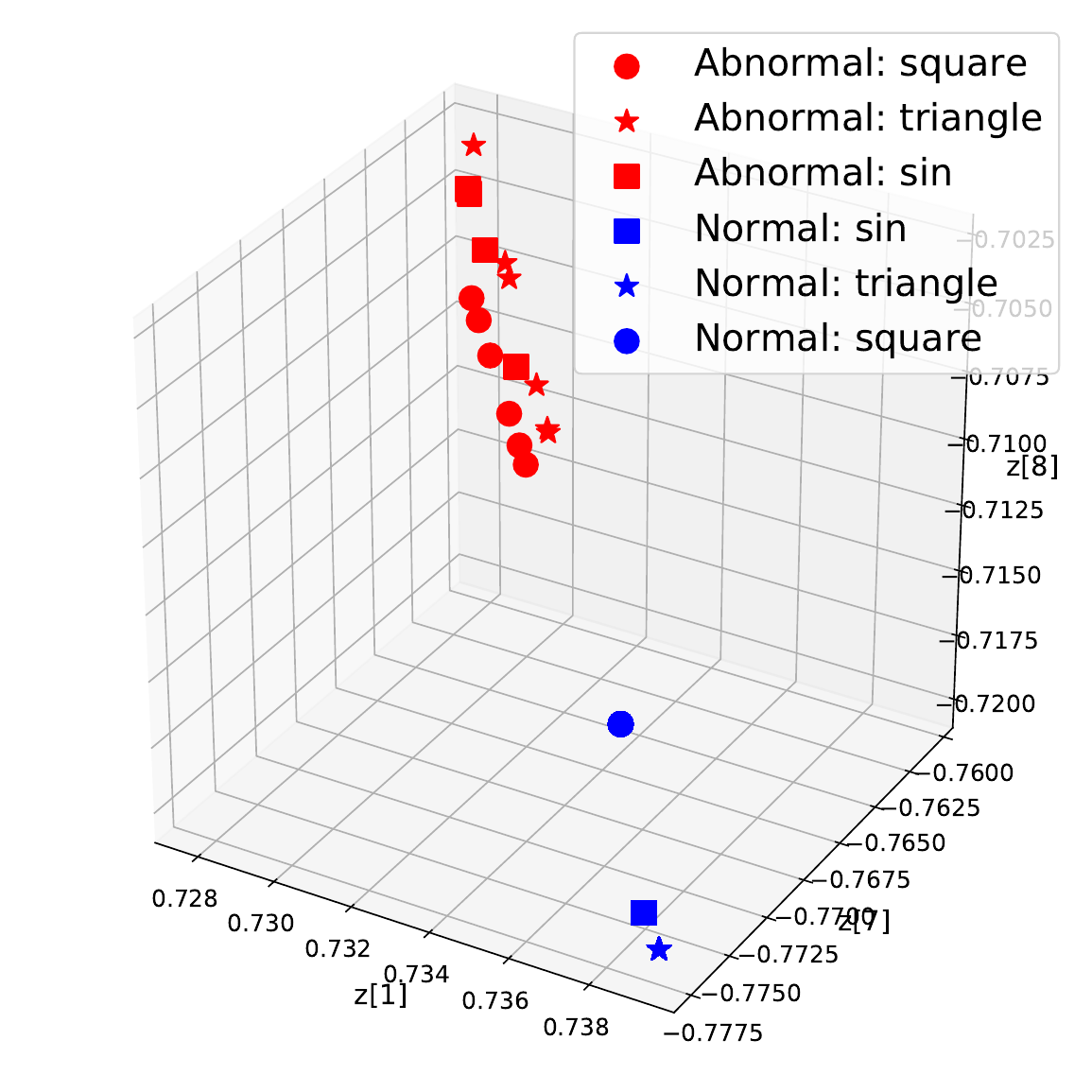}
        \end{minipage}%
    }
    \caption{\textcolor{black}{Exemplary samples, 2D and 3D latent space visualizations of type-1 to type-4 anomalies. An SCR greater than 1, with higher values, indicates better visualization performance.}}
    \label{fig:result_anomaly}
\end{figure*}

A total of four types of anomalies\footnote{\url{https://anomaly.io/anomaly-detection-twitter-r/}} have been generated to assess the effectiveness of the proposed ET-Net framework. 
Type-1 anomaly refers to strong local additive white Gaussian noise, as shown in Figure \ref{fig:result_anomaly}(a1).
Type-2 anomaly stands for an unusually high activity that spans a short period of time (Figure \ref{fig:result_anomaly}(b1)).
Type-3 is the ``breakdown'' anomaly (Figure \ref{fig:result_anomaly}(c1)).
Type-4 anomaly is created by adding an impulse noise into the time series, as shown in Figure \ref{fig:result_anomaly}(d1).
Both the time domain and latent space representations are illustrated in Figure \ref{fig:result_anomaly}. In the second and third sub-figures of Figure \ref{fig:result_anomaly}, the blue symbols represent non-anomalous time series while the red ones correspond to anomaly time series. It is seen through both 2D and 3D visualization that ET-Net can effectively separate the anomalous time series from non-anomalous ones in the latent space.

\begin{figure}[!tbp]
\centering
    \includegraphics[width=0.6\columnwidth]{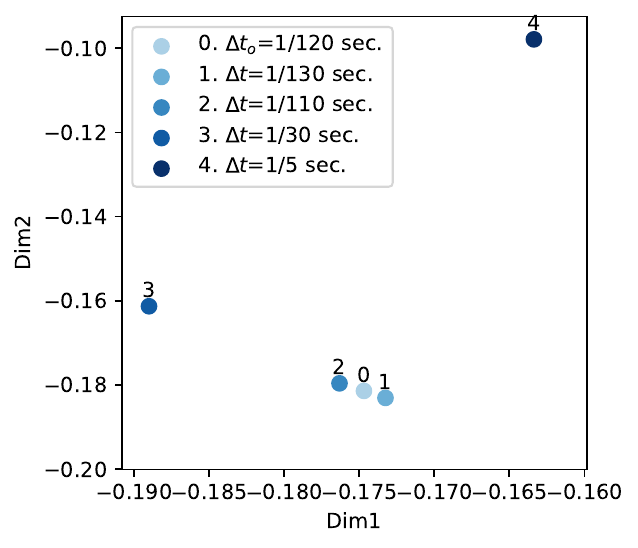}
    \caption{Latent space representation of sine signals with different sampling intervals.}
\label{fig:latent_sin_diff}
\end{figure}

\begin{figure*}
    \centering
    \subfloat[Type-1 Anomaly, SCR of 2D visualization is 11.28]{%
        \begin{minipage}{0.48\textwidth}
            \centering
            \includegraphics[width=0.48\linewidth]{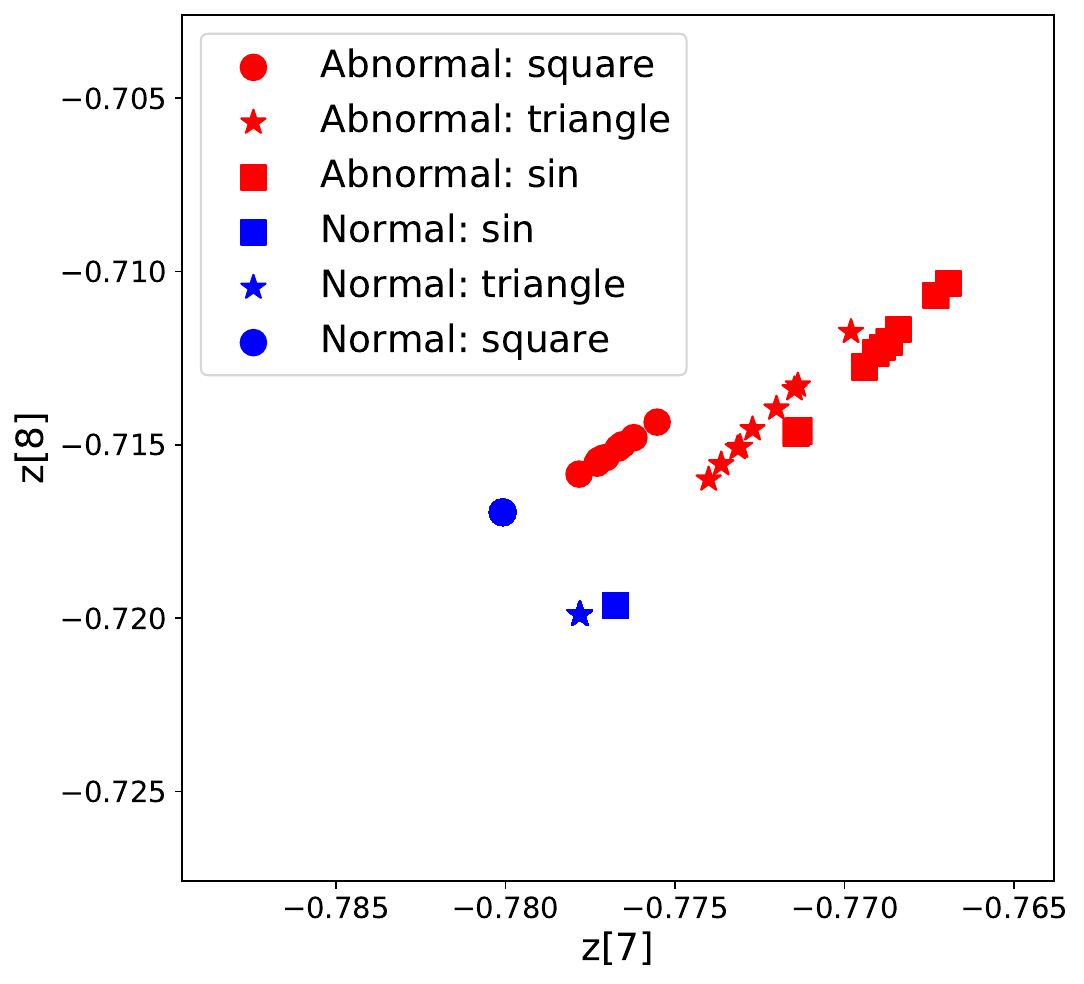}\hfil
            \includegraphics[width=0.48\linewidth]{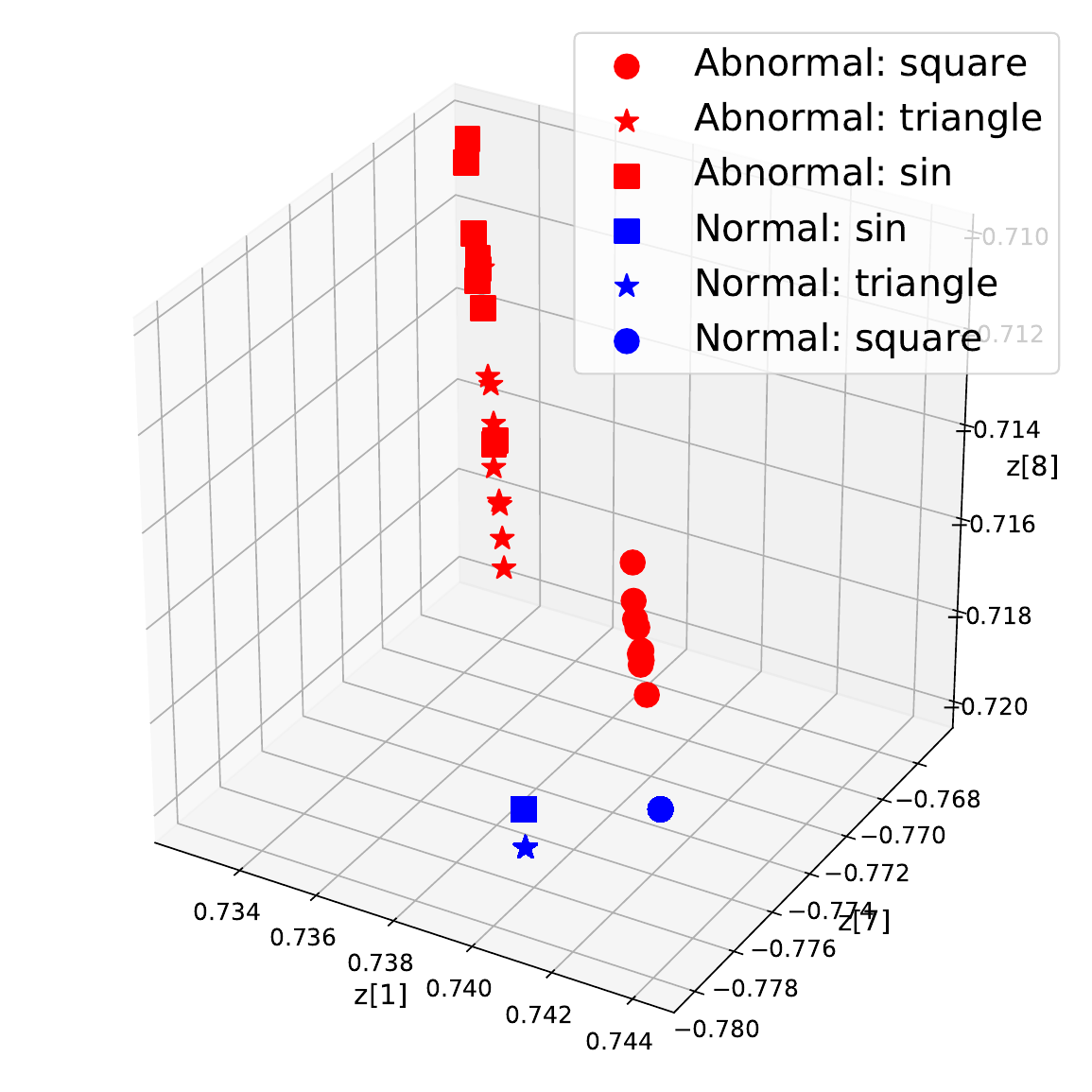}
        \end{minipage}%
    }\hfil
    \subfloat[Type-2 Anomaly, SCR of 2D visualization is 18.90]{%
        \begin{minipage}{0.48\textwidth}
            \centering
            \includegraphics[width=0.48\linewidth]{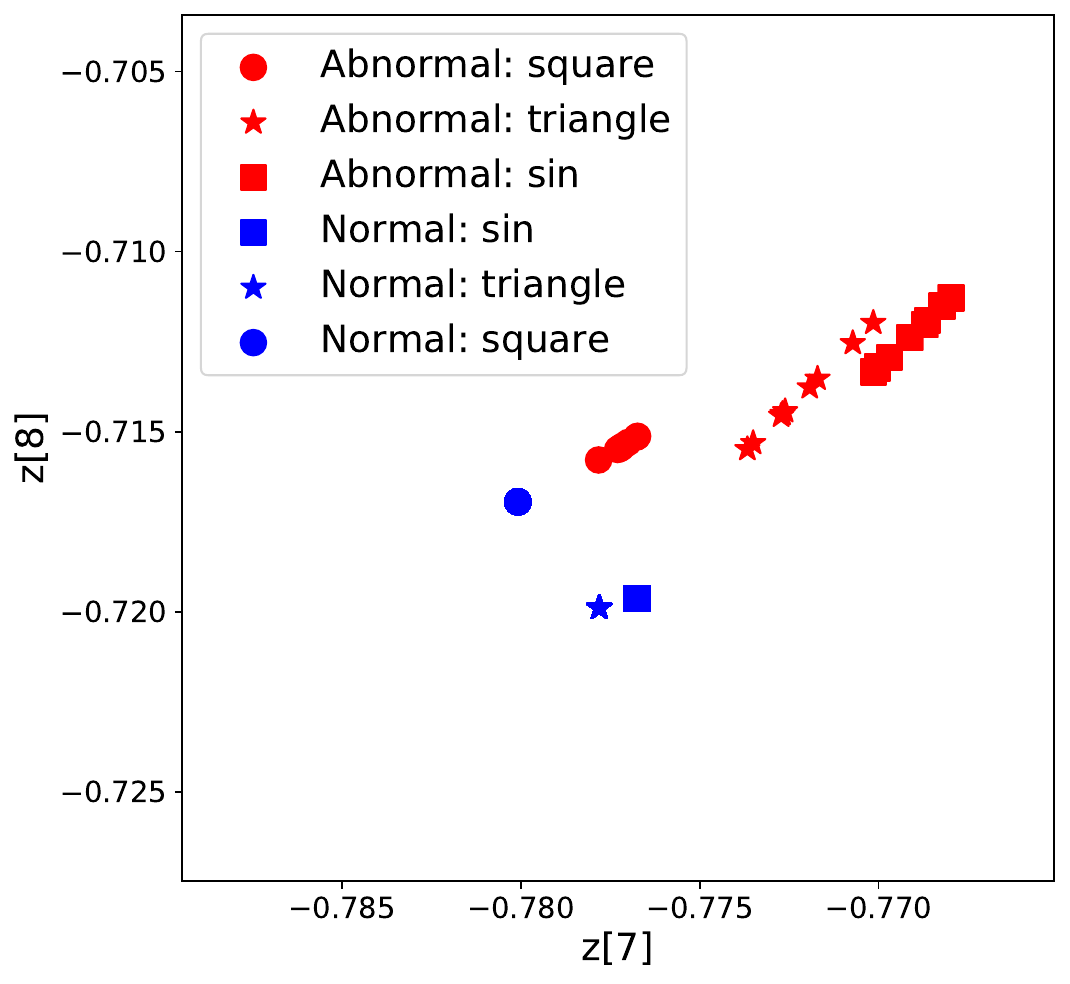}\hfil
            \includegraphics[width=0.48\linewidth]{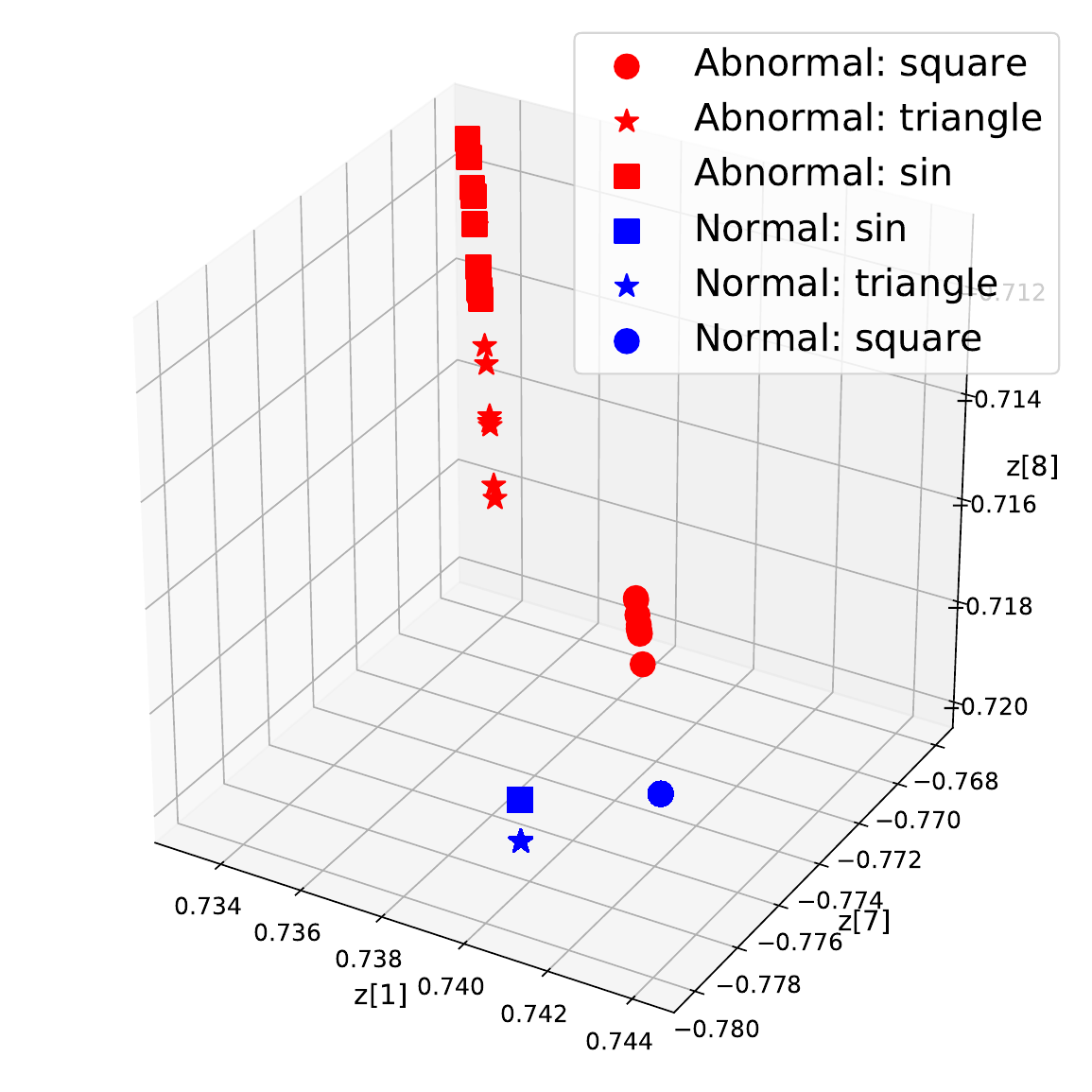}
        \end{minipage}%
    } \\ 

    \subfloat[Type-3 Anomaly, SCR of 2D visualization is 6.35]{%
        \begin{minipage}{0.48\textwidth}
            \centering
            \includegraphics[width=0.48\linewidth]{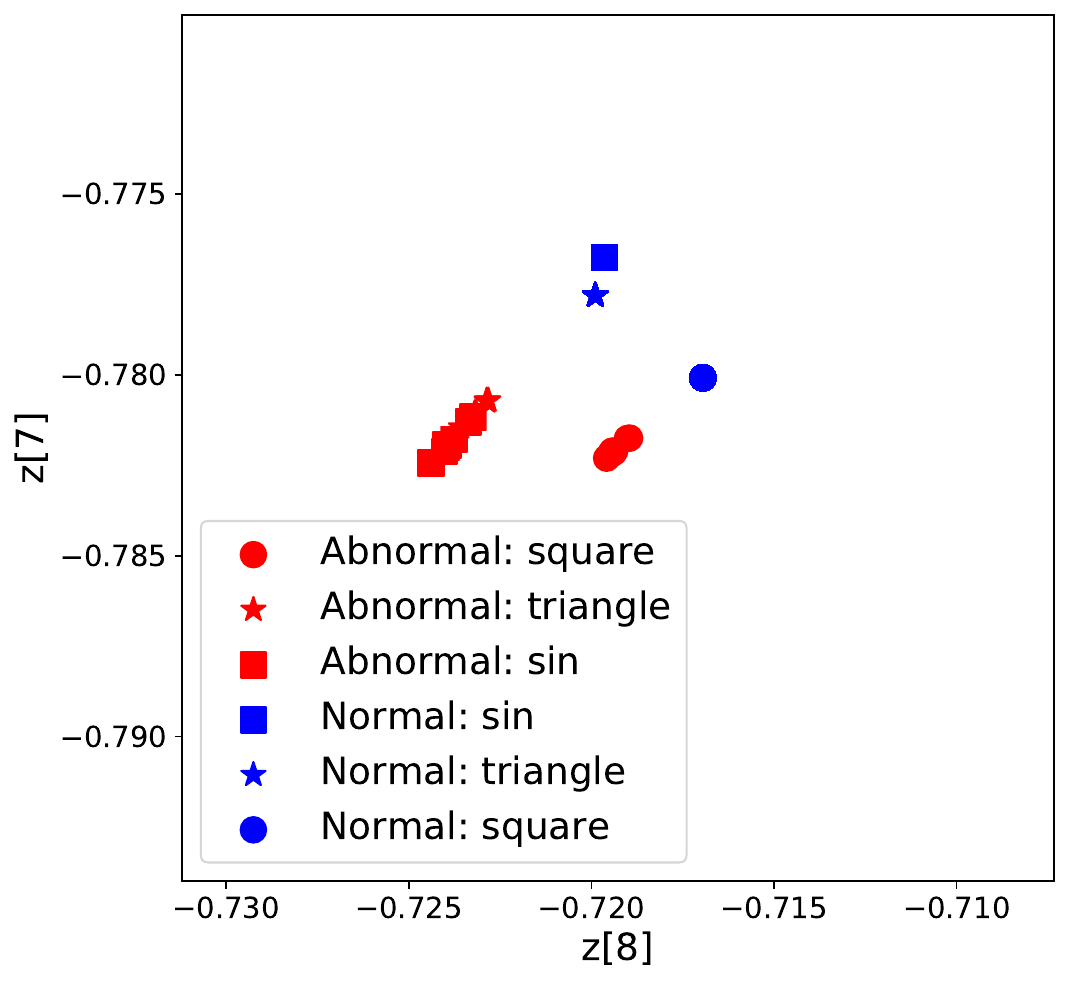}\hfil
            \includegraphics[width=0.48\linewidth]{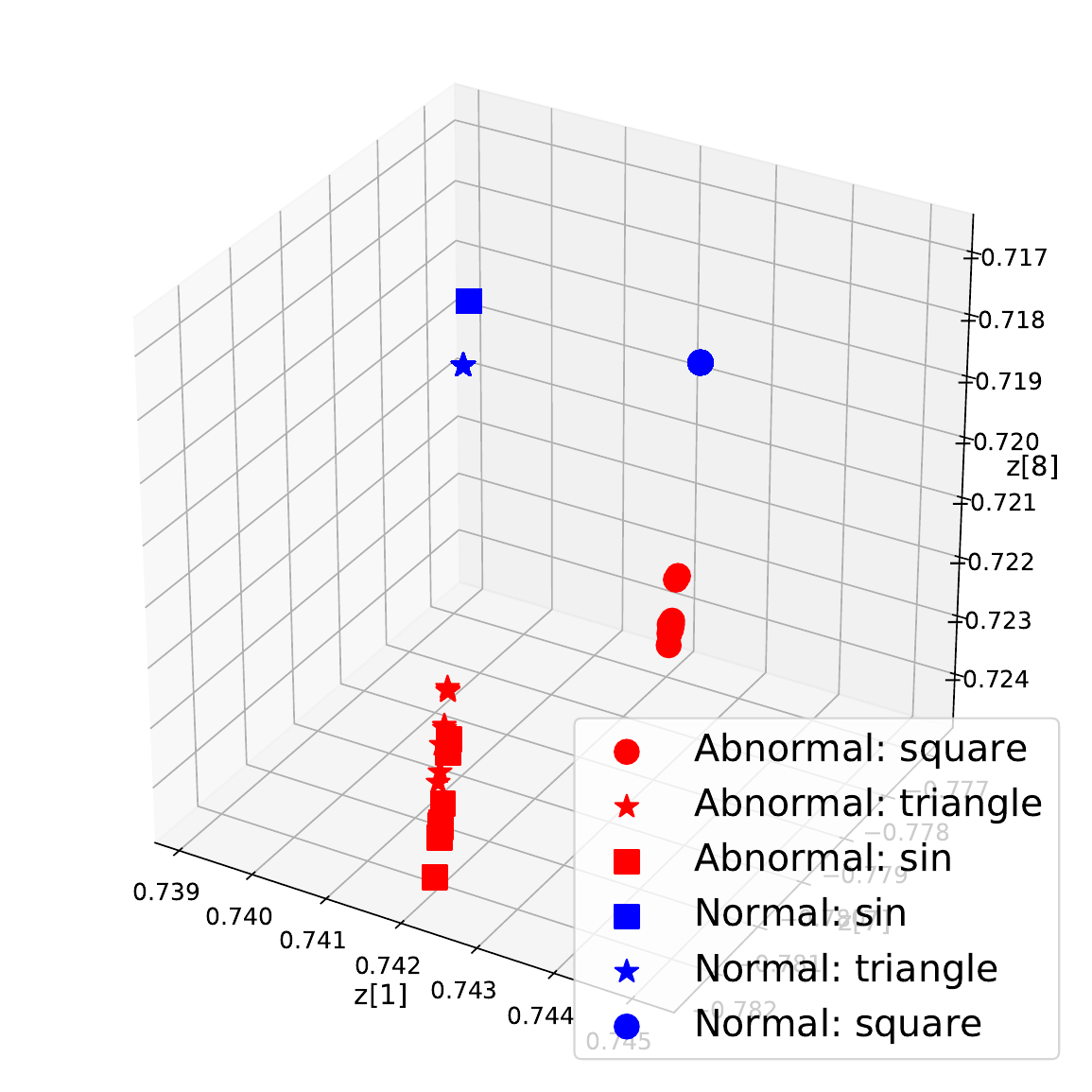}
        \end{minipage}%
    }\hfil
    \subfloat[Type-4 Anomaly, SCR of 2D visualization is 18.61]{%
        \begin{minipage}{0.48\textwidth}
            \centering
            \includegraphics[width=0.48\linewidth]{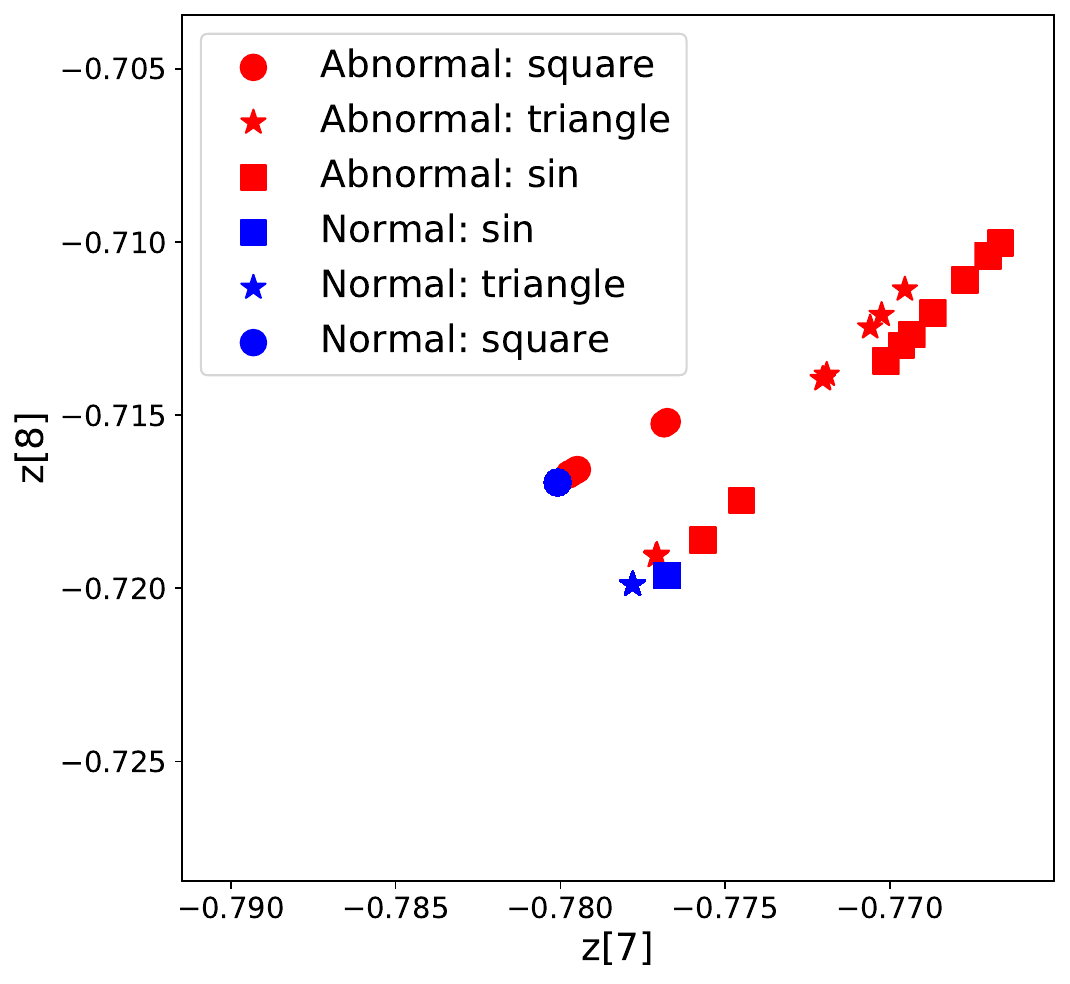}\hfil
            \includegraphics[width=0.48\linewidth]{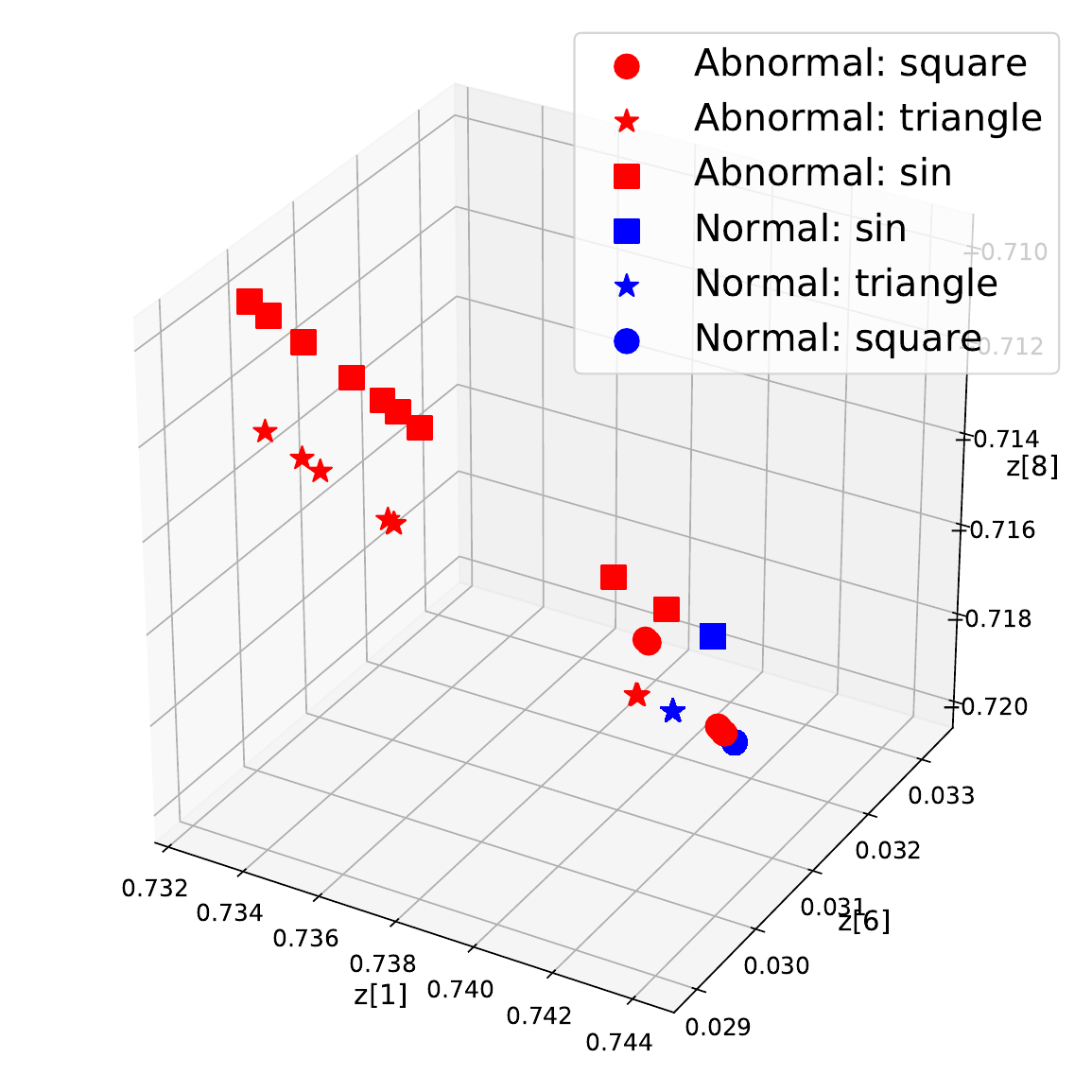}
        \end{minipage}%
    }
    \caption{\textcolor{black}{2D and 3D latent space representations of type-1 to type-4 anomalies with different sampling intervals. An SCR greater than 1, with higher values, indicates better visualization performance.}}
    \label{fig:c_result_anomaly}
\end{figure*}

\textbf{Robustness against time granularity variations:} 
We then explored the ability bounds of ET-Net to generate robust representations when the data granularity changes, taking a sin signal with a frequency of 10 Hz as an example, and Figure \ref{fig:exp_anomaly}(a) is an example of one of the cycles. We set the original sampling interval $\Delta t_o$ to $1/120$ second, and then vary the sampling rates from $\Delta t = 1/110, 1/130$ second to $\Delta t = 1/30, 1/5$ second, and visualize the obtained latent space representations in Figure \ref{fig:latent_sin_diff}. Specifically, ET-Net produced robust time series representations when the data had small sampling frequency variations relative to sample 0 (e.g., 130 Hz,110 Hz for samples 1 and 2). And when the sampling frequency is close to the Nyquist sampling frequency (20Hz) for sample 3, compared to sample 4, which does not satisfy the sampling law, ET-Net generates a relatively more robust representation, as shown by point 3 being closer to point 0,1 and 2 while point 4 being far from all data points.
Next, we apply the obtained anomlay detection model to a test time series dataset in which the sampling rate differs from that of the training dataset.
Figure \ref{fig:c_result_anomaly} illustrates the latent space representations of type-1 to type-4 anomaly time series and corresponding non-anomalous time series, where red and blue symbols represent abnormal and normal time series, respectively. It is seen that a ET-Net can be applied to time series with a different time granularity without any model retraining.

\subsubsection{Clustering}
\begin{figure}[!tbp]
\centering
    \subfloat[Sine waves]{
    \begin{minipage}{0.32\columnwidth}
        \includegraphics[width=\columnwidth]{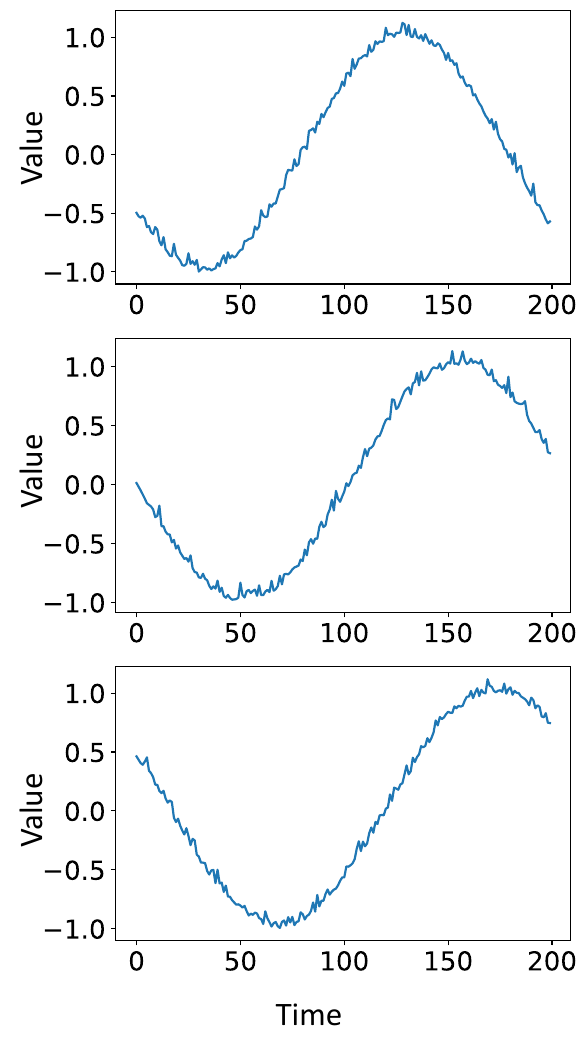}
    \end{minipage}}
    \subfloat[Square waves]{
    \begin{minipage}{0.32\columnwidth}
        \includegraphics[width=\columnwidth]{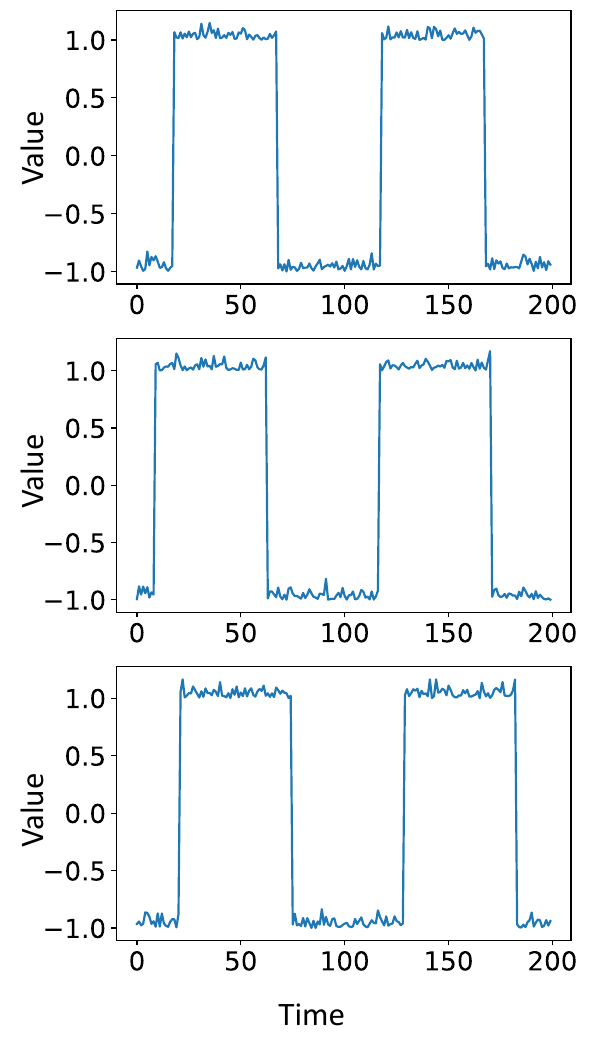}
    \end{minipage}}
    \subfloat[Triangle waves]{
    \begin{minipage}{0.32\columnwidth}
        \includegraphics[width=\columnwidth]{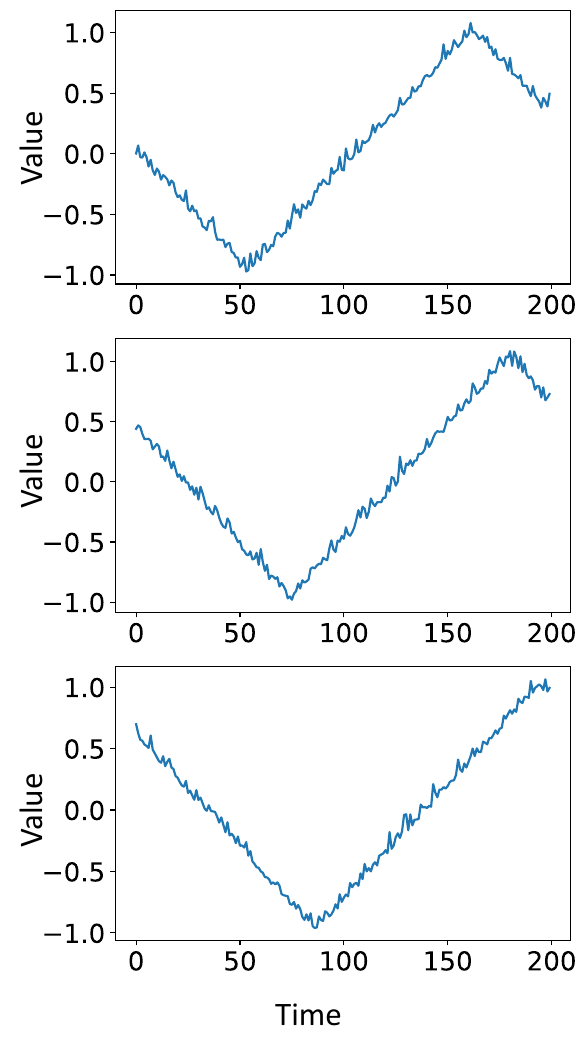}
    \end{minipage}}
    \caption{\textcolor{black}{Three category of
    samples from the synthetic clustering dataset.}}
\label{fig:exp_clustering}
\end{figure}

\begin{figure}[!tbp]
\centering
    \subfloat[2D visualization, SC is 0.60]{
    \begin{minipage}[!t]{0.48\columnwidth}
        \includegraphics[width=\columnwidth]{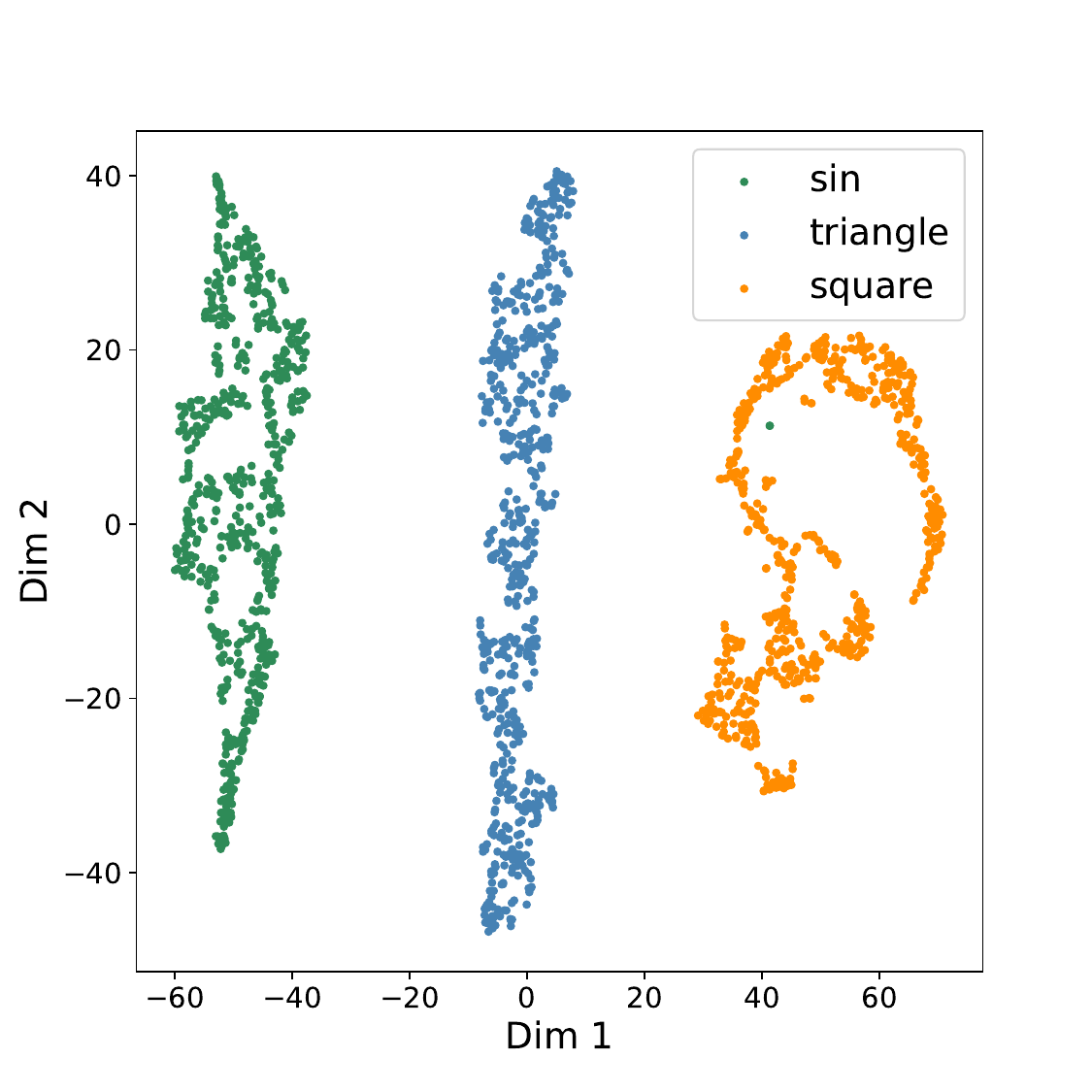}
    \end{minipage}}
    \subfloat[3D visualization, SC is 0.43]{
    \begin{minipage}[!t]{0.48\columnwidth}
        \includegraphics[width=\columnwidth]{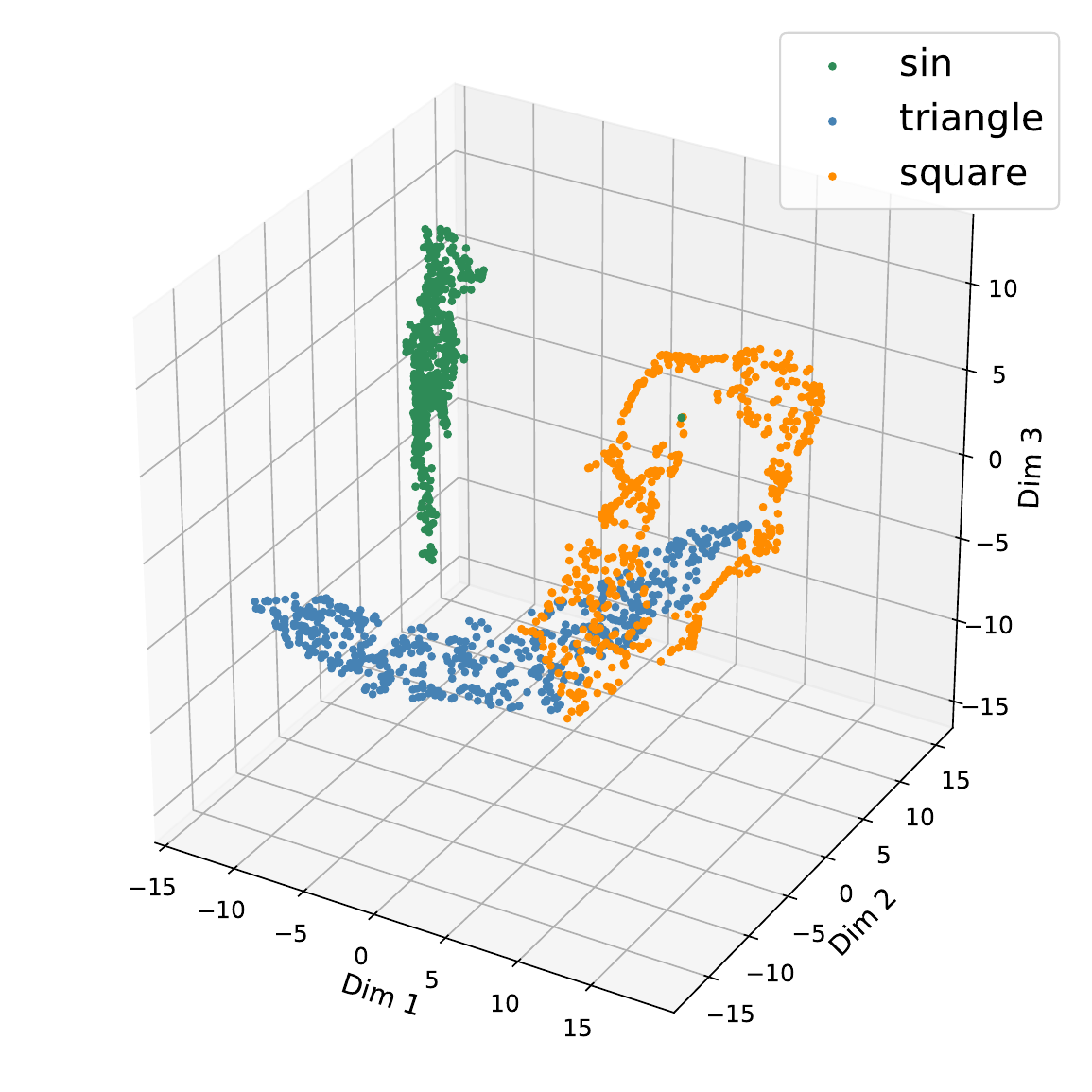}
    \end{minipage}}
    \caption{2D and 3D visualization of clustering. An SC closer to 1 indicates better performance, whereas an SC approaching -1 signifies poorer clustering.}
\label{fig:result_clustering}
\end{figure}

This dataset consists of three types of time series, i.e., sine waves, square waves and triangle waves (as shown in different columns of Figure \ref{fig:exp_clustering}), we also pass these time series through an Additive white Gaussian noise (AWGN) channel and introduce phase difference artificially (as shown in different rows of Figure \ref{fig:exp_clustering}) to make them more realistic. The vector embeddings in the latent space can be obtained for the time series through the ET-Net, as visualized in Figure \ref{fig:result_clustering} using the t-SNE algorithm. It is evident that we can easily cluster these time series in the latent space.

\begin{figure}[!tbp]
\centering
    \subfloat[Type-1 noise]{
        \begin{minipage}{0.475\columnwidth}
        \includegraphics[width=\columnwidth]{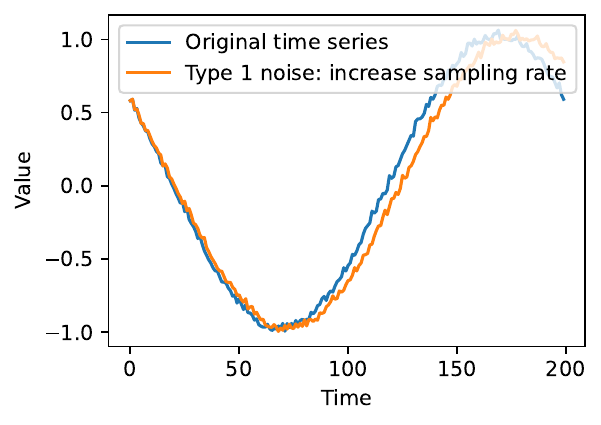}
        \end{minipage}}
    \subfloat[Type-2 noise]{
        \begin{minipage}{0.475\columnwidth}
        \includegraphics[width=\columnwidth]{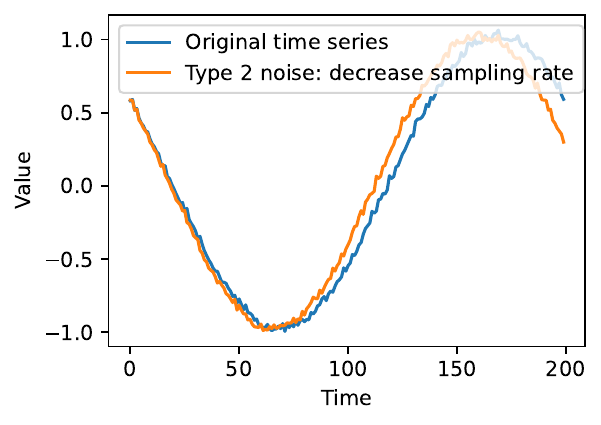}
        \end{minipage}}

    \subfloat[Type-3 noise]{
        \begin{minipage}{0.475\columnwidth}
        \includegraphics[width=\columnwidth]{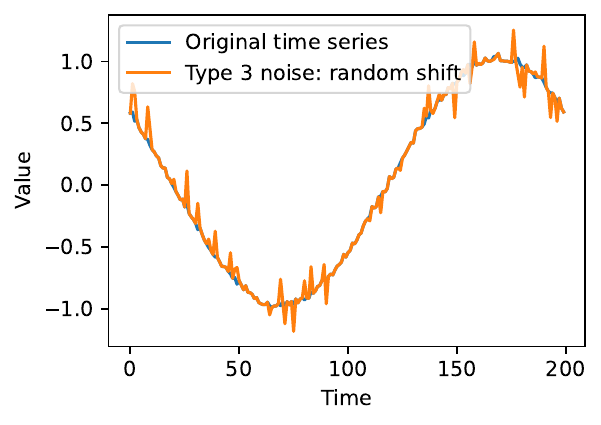}
        \end{minipage}}
    \subfloat[Type-4 noise]{
        \begin{minipage}{0.475\columnwidth}
        \includegraphics[width=\columnwidth]{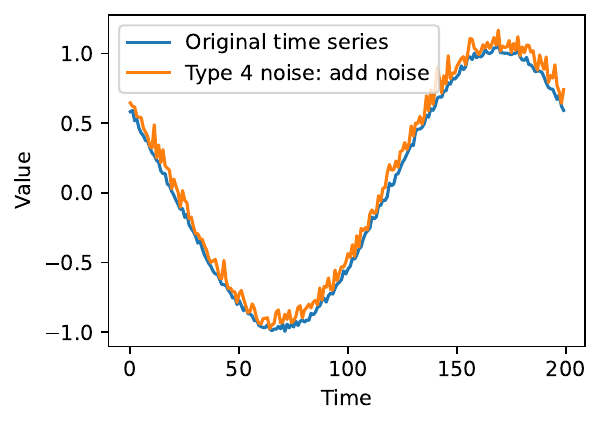}
        \end{minipage}}
\caption{\textcolor{black}{Exemplary synthetic samples of four types of noise, using a sine signal as an example.}}
\label{fig:exp_noise}
\end{figure}

\textbf{Robustness against different types of noise:}
Four types of noise described in \cite{wang2013effectiveness} are considered in this experiment.
Type-1 and type-2 noise stand for increasing (Figure \ref{fig:exp_noise}(a)) and decreasing sampling rate (Figure \ref{fig:exp_noise}(b)), respectively.
Type-3 noise is the shifting noise (Figure \ref{fig:exp_noise}(c)).
Type-4 noise refers to adding Gaussian noise to the entire time series (Figure \ref{fig:exp_noise}(d)).

\begin{figure}[!tbp]
\centering
    \includegraphics[width=\columnwidth]{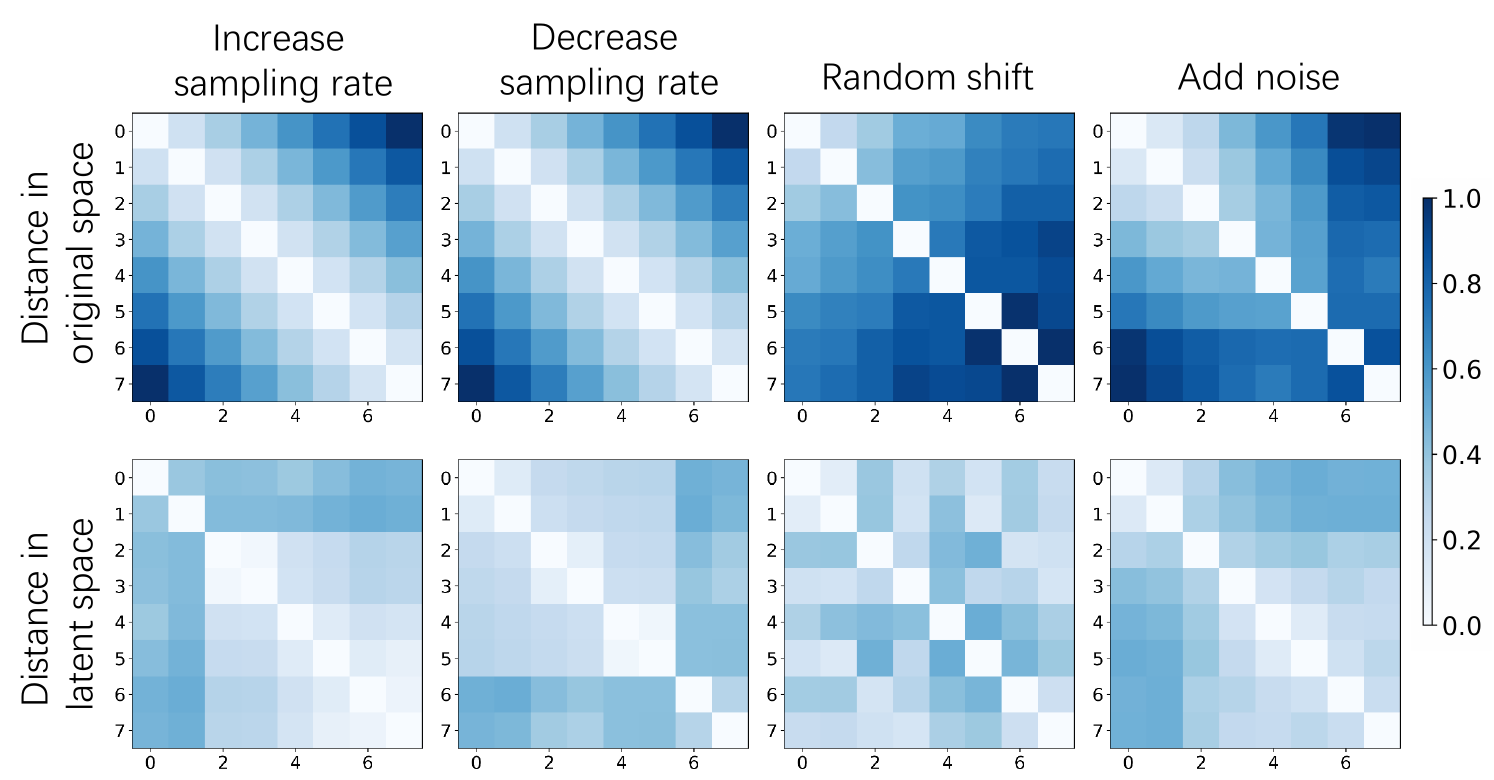}
    \caption{Normalized Euclidean distance matrix of a collection of time series in the original space (top row) and latent space (bottom row). The four columns are distance matrices when the time series contains different types of noise. }
\label{fig:result_noise}
\end{figure}

We then apply the four types of noise to a sine time series and compute the Euclidean distance between the original time series and the time series with noise in both original and latent spaces.
As shown in Figure \ref{fig:result_noise}, for all four types of noise, the Euclidean distance will increase quickly with the level of noise.
In contrast, the proposed framework remains effective in the presence of all four types of noise and can mine the similarity between the original time series and the ones with noise.
As a matter of fact, as shown in Table \ref{table:noise}, it is the only method that remains robust against all types of noise.

\begin{table}[!tbp]
\centering
\caption{Comparison of time series similarity measures \cite{wang2013effectiveness,magdy2015review}}
\label{table:noise}
\scalebox{1}{
\begin{tabular}{@{}ccccc@{}}
    \toprule
                           & ED        & DTW       & EDR       & ET-Net \\ \midrule
    \makecell[c]{Increase\\sampling\\rate} & Sensitive & Fair      & Sensitive & Robust      \\
    \makecell[c]{Decrease\\sampling\\rate} & Sensitive & Sensitive & Fair      & Robust      \\
    \makecell[c]{Random\\shift}           & Robust    & Robust    & Robust    & Robust      \\
    \makecell[c]{Add\\noise}              & Sensitive & Sensitive & Robust    & Robust      \\ \bottomrule
\end{tabular}}
\end{table}

\subsection{Anomaly Detection Results on Real-world Datasets}
\label{res_anomaly_dectection_real_world}

\subsubsection{Implementation Details}
We conduct experiments on a workstation with 2 NVIDIA Titan V GPUs. TensorFlow is employed to implement the proposed ET-Net framework.

ET-Net contains four types of hyperparameters,
1) the number of encoders or decoders $N_E$ in the W compression network;
2) the number of layers $N_L$ in the D compression network;
3) the number of neurons $N_N$ in each encoder or decoder;
4) the number of mixture components $K$ in GMM.
We perform a grid search to determine the hyperparameters of ET-Net. The hyperparameter settings for ET-Net are listed in Table \ref{table:hyper_params}.
LSTM cell and GRU cell are adopted in the W and D compression network respectively. Please note that either GRU or LSTM can be used as the recurrent unit for both compression networks, in most cases,  we recommend using GRU as the recurrent unit, considering the trade-off between model size and performance. We use the Adam optimization algorithm \cite{kingma2014adam} to train the proposed model, and the initial learning rate is set to $10^{-3}$.

\begin{table}[!tbp]
\centering
\caption{Hyperparameter settings}
\label{table:hyper_params}
\scalebox{1}{
    \begin{tabular}{@{}lcccc@{}}
    \toprule
    Dataset      & $N_E$ & $N_L$ & $N_N$ & $K$ \\ \midrule
    \multicolumn{5}{c}{Anomaly Detection} \\ \midrule
    UNSW-IoT       & 3              & 2            & 18            & 4                    \\
    IoT23          & 3              & 2            & 18            & 4                    \\
    Cell traffic   & 3              & 4            & 18            & 1                    \\
    MedicalImages  & 3              & 2            & 8             & 3                    \\
    MoteStrain     & 3              & 2            & 8             & 1                    \\
    PowerCons      & 3              & 2            & 12            & 1                    \\
    SmoothSubspace & 3              & 2            & 8             & 1                    \\ \midrule
    \multicolumn{5}{c}{Clustering} \\ \midrule
    UNSW-IoT     & 3              & 2            & 12            & 3                    \\
    IoT23        & 3              & 4            & 18            & 8                    \\
    Cell traffic & 3              & 2            & 12            & 4                    \\ \bottomrule
    \end{tabular}}
\end{table}

\subsubsection{Effectiveness}
The performance of ET-Net and other state-of-the-art methods on a total of seven real-world datasets are listed in Table \ref{table:auc}. ET-Net-W and ET-Net-D represent W compression network with GMM and D compression network with GMM, respectively. It is evident that the proposed ET-Net outperforms other competing methods considerably. It ranks first in five out of seven datasets, and ranks second in the remaining two datasets. In particular, for the cell traffic datasets, ET-Net outperforms the second-best method by around 14\%. Furthermore, ET-Net outperforms ET-Net-W and ET-Net-D thanks to the ensemble of multiple networks.
For the MoteStrain dataset, shared-SRNN and GRU-AE, based only on reconstruction errors, show superior detection performance compared to ET-Net-W and ET-Net-D, which integrate data representations with reconstruction errors. This is attributed to the severe shortage of training samples, that is, only 10 training samples, leading to the undertraining of large-parameter models, hindering the effective differentiation of normal and abnormal latent representations.

% \usepackage{booktabs}
\begin{table*}[]
\centering
\caption{Anomaly detection performance measured by AUC. Optimal and suboptimal results are bolded and underlined, respectively.}
\label{table:auc}

\begin{tabular}{@{}lccccccc@{}}
\toprule
Method          & UNSW-IoT        & Cell traffic    & loT23           & Medicallmages   & MoteStrain      & PowerCons       & SmoothSubspace  \\ \midrule
OCSVM           & 0.7947          & 0.3241          & 0.5000          & 0.5366          & 0.8017          & 0.9827          & 0.9920          \\
LoF             & 0.8705          & 0.5940          & 0.6402          & 0.6689          & 0.5339          & 0.7630          & 0.9200          \\
IF              & 0.8586          & 0.4758          & 0.6544          & 0.5012          & 0.8959          & 0.9679          & 0.9920          \\
DTW             & 0.8413          & 0.4865          & 0.6480          & 0.4508          & 0.8550          & 0.9642          & \textbf{1.0000} \\ \midrule
KitNET          & 0.9196          & 0.1667          & 0.2879          & 0.5074          & 0.7112          & \textbf{0.9975} & \textbf{1.0000} \\
GRU-AE          & 0.9099          & 0.4259          & 0.7430          & 0.6358          & 0.9122          & 0.9691          & 0.9840          \\
Shared-SRNN     & 0.8279          & \underline{0.6944}          & 0.7657          & 0.4680          & \textbf{0.9397} & 0.9716          & 0.9720          \\
DAGMM           & 0.8314          & 0.5000          & 0.7964          & 0.6473          & 0.5000          & 0.5333          & 0.7800          \\
USAD            & \underline{0.9384}          & 0.6482          & 0.3152          & 0.5782          & 0.8253          & \underline{0.9840}          & \underline{0.9960}          \\ \midrule
ET-Net-W        & 0.9356          & 0.3982          & \textbf{0.8504} & \underline{0.7584}          & 0.7597          & 0.9382          & 0.9840          \\
ET-Net-D        & 0.8696          & \textbf{0.8333} & 0.6819          & 0.5724          & 0.9016          & 0.9506          & \textbf{1.0000} \\
\textbf{ET-Net} & \textbf{0.9503} & \textbf{0.8333} & \underline{0.8289}          & \textbf{0.7608} & \underline{0.9270}          & \textbf{0.9975} & \textbf{1.0000} \\ \bottomrule
\end{tabular}
\end{table*}

In the previous study, we assume all the training dataset constitutes non-anomalous time series. However, such an assumption does not always hold in practice, since a small portion of the training data might be anomalies.
As a remedy, we artificially inject anomalies into the training dataset to check whether we can still obtain an effective anomaly detector. Table \ref{table:robustness} demonstrates that the proposed ET-Net architecture remains effective even in the presence 10\% of anomalies in the training dataset.

\begin{table}[!tbp]
\caption{AUC with injected anomalies in training set}
\label{table:robustness}
\centering
\scalebox{1}{
    \begin{tabular}{@{}cclcc@{}}
    \multicolumn{2}{c}{UNSW-IoT} & \multicolumn{1}{c}{} & \multicolumn{2}{c}{Cell traffic} \\ \cmidrule(r){1-2} \cmidrule(l){4-5}
    Proportions & AUC Score &  & Proportions & AUC Score \\ \cmidrule(r){1-2} \cmidrule(l){4-5}
    0\% & 0.9503 &  & 0\% & 0.8333 \\
    5\% & 0.9177 &  & 5\% & 0.8229 \\
    10\% & 0.9071 &  & 10\% & 0.7779 \\ \cmidrule(r){1-2} \cmidrule(l){4-5}
    \textbf{loss} & \textbf{4.32\%} &  & \textbf{loss} & \textbf{5.54\%} \\ \cmidrule(r){1-2} \cmidrule(l){4-5}
    \end{tabular}}
\end{table}

\subsubsection{Robustness against Data Granularity and Traffic Disturbances}
\begin{figure}[!tbp]
\centering
    \includegraphics[width=0.8\columnwidth]{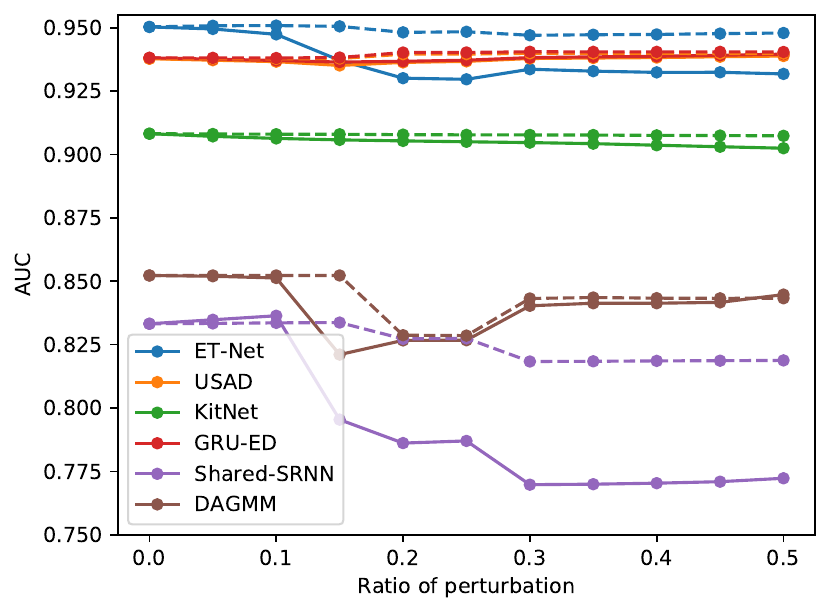}
    \caption{AUC with perturbation in the testing set, where the solid and dashed lines represent adding adversarial perturbations and dummy data packets, respectively.}
\label{fig:adc_auc}
\end{figure}

Time series generated by event-triggered sensors can be highly complex due to different sampling intervals. This raises the question of whether an ET-Net model trained on time series with one sampling interval can be applied to a time series with other intervals. To investigate this, we trained an ET-Net model on a dataset with a 60-second sampling interval and tested its performance on time series with sampling intervals varying from 60 to 120 seconds. Table \ref{table:sample_rate} shows that the model remains effective even when the data granularity changes, indicating its robustness.
We also tested the robustness of ET-Net to traffic disturbances that may occur in real-world scenarios. We applied the trained model to a test dataset with different perturbation ratios, the injected disturbances include adding dummy data packets \cite{cai2014systematic} and adversarial perturbations \cite{nasr2021defeating}. Figure \ref{fig:adc_auc} shows that even in the worst case where half of the samples are perturbed, the AUC of ET-Net is only reduced by about 2\%, which is still better than the majority of baselines, demonstrating its robustness.

\begin{table}[!tbp]
\centering
\caption{AUC on different sampling intervals. Optimal and suboptimal results are bolded and underlined, respectively.}
\label{table:sample_rate}
\scalebox{1}{
    \begin{tabular}{@{}lcccc@{}}
    \toprule
    Sampling Intervals & 60sec           & 90sec           & 120sec          \\ \midrule
    OCSVM              & 0.7947          & 0.5010          & 0.5000          \\
    LoF                & 0.8705          & 0.8780          & 0.8664          \\
    IF   & 0.8586          & 0.8211          & 0.7939          \\
    DTW                & 0.8413          & 0.8511          & 0.8350          \\ \midrule
    KitNET             & 0.9196    & 0.8501   & 0.8143    \\
    GRU-AE             & 0.9099    & \underline{0.8903}    & 0.8792    \\
    Shared-SRNN       & 0.8279          & 0.7541          & 0.7317          \\
    DAGMM              & 0.8314          & 0.7591          & 0.7494          \\ 
    USAD              & \underline{0.9384}          & 0.8803          & \underline{0.8930}          \\ \midrule
    \textbf{ET-Net}             & \textbf{0.9503} & \textbf{0.8909} & \textbf{0.9078} \\ \bottomrule
    \end{tabular}}
\end{table}

\subsubsection{Visualization and Interpretability}
\label{sec:inter}
From a perspective of latent space visualization, ET-Net models the distribution of normal samples using a GMM model, and forms a normal cluster in the latent space in the anomaly detection task. Thus, the vector embedding that deviates from this distribution is deemed as an anomaly. A typical example is shown in Figure \ref{fig:result_anomaly}.

\begin{figure}[!tbp]
\centering
    \includegraphics[width=1.05\columnwidth]{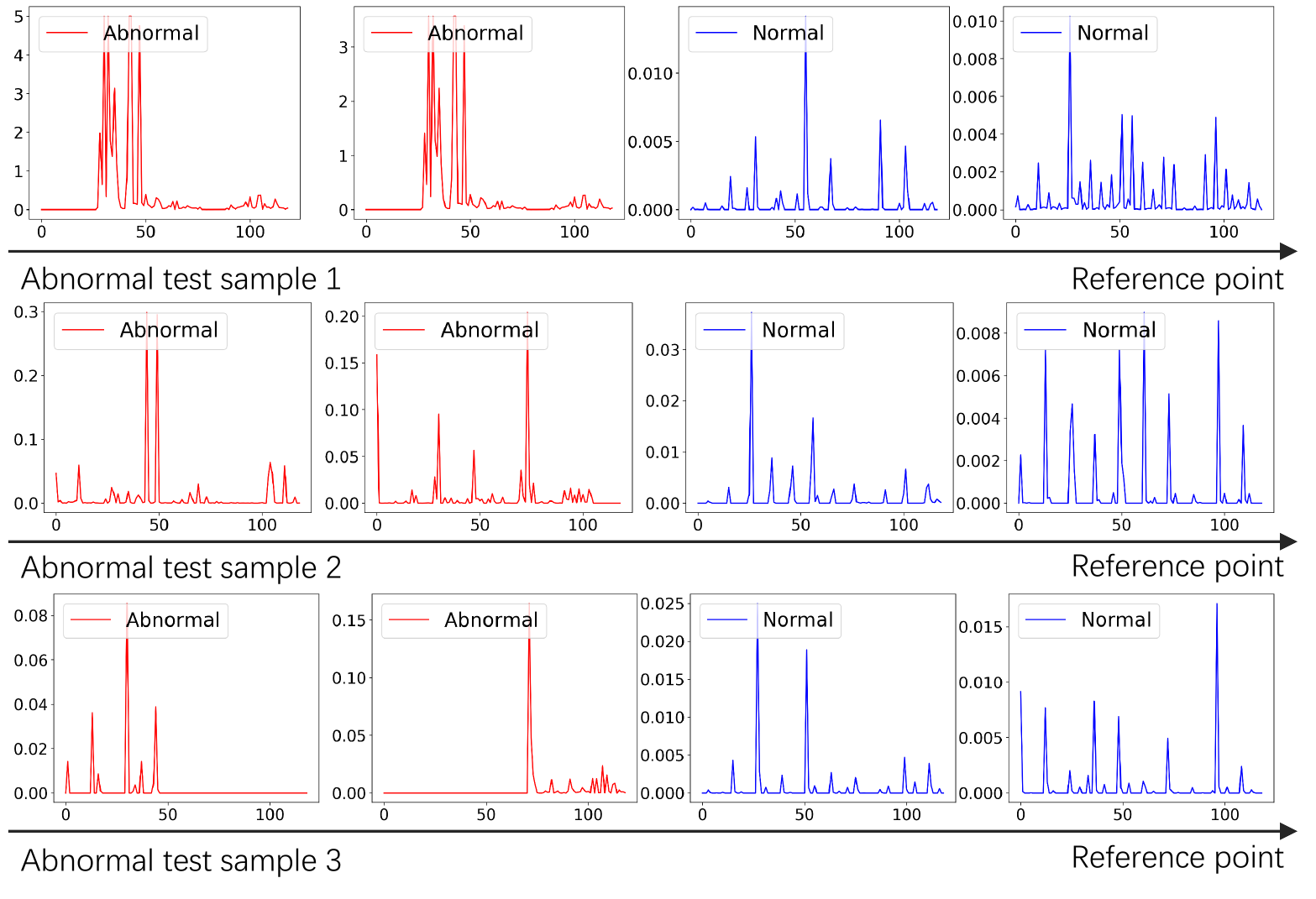}
    \caption{Exemplary test samples and corresponding reference samples from the UNSW-IoT dataset. \textcolor{black}{The x-axis represents time in minutes and the y-axis represents traffic in kilobytes.}}
\label{fig:visual_ts}
\end{figure}

\begin{figure*}[!t]
\centering
    \includegraphics[width=1.8\columnwidth]{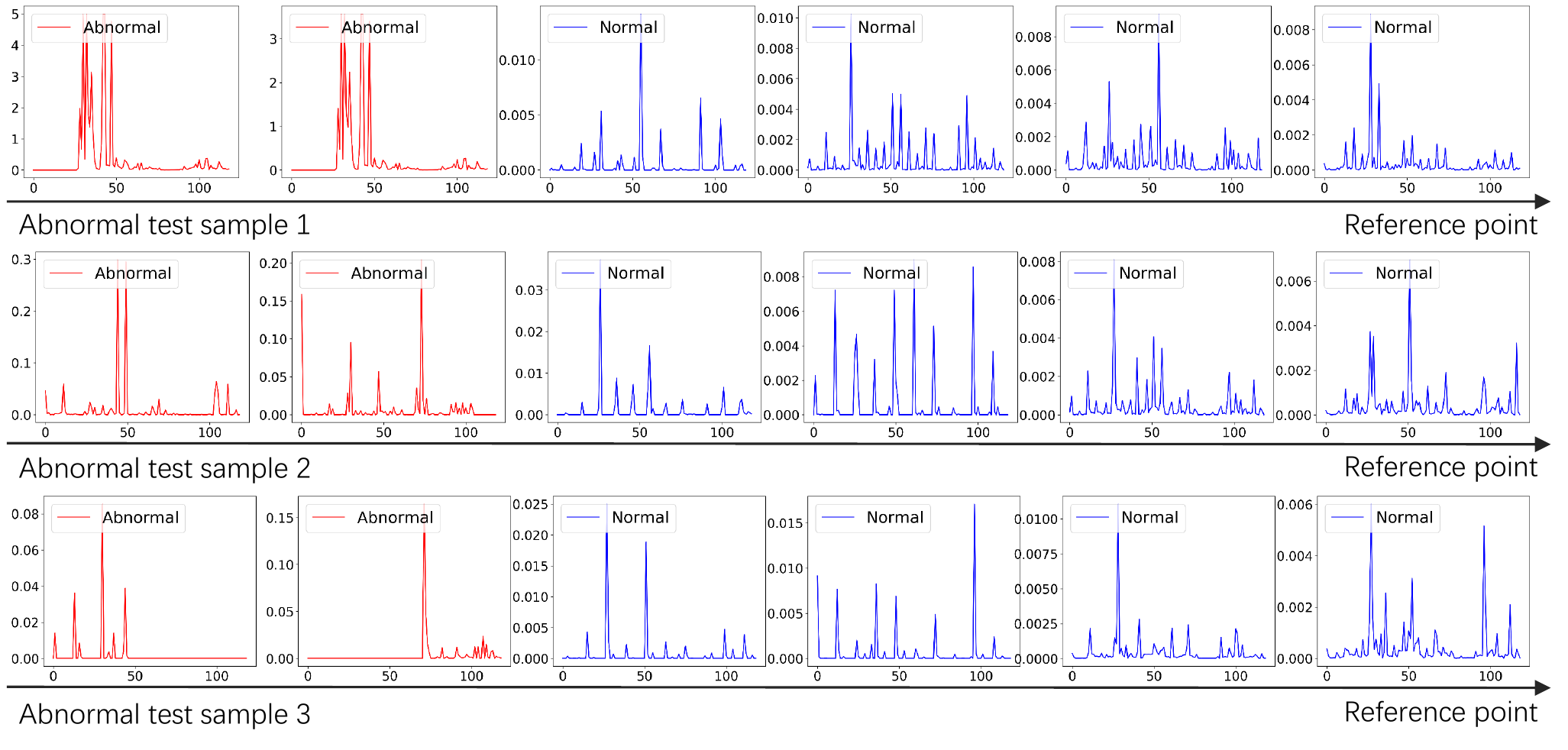}
    \caption{Exemplary test samples and corresponding reference samples from the UNSW-IoT dataset. The first column is the test samples, and each subsequent column is closer to the center of the normal cluster. \textcolor{black}{The x-axis represents time in minutes and the y-axis represents traffic in kilobytes.}}
\label{fig:visual_ts_complete}
\end{figure*}

Based on the fact that normal samples are grouped into clusters in the latent space, we propose an \textit{example-based attribution method} to explain the detected anomalies. Specifically, given a time series $\mathbf{x}_a$ that is deemed as an anomaly, we draw a straight line in latent space from the representation of $\mathbf{x}_a$ to the center of normal cluster $\mathbf{z}_{cnt}$. We call this line reference line hereafter. The comparison among the anomaly time series and corresponding reference time series around the reference line helps to explain the difference between the anomalous times series and the normal ones. Note that the reference samples are selected from the training set.
Figures in the first column in Figure \ref{fig:visual_ts} illustrate three representative abnormal time series, and the remains are reference time series, where the samples in the second column are the reference sample closest to the abnormal samples, and the third and fourth columns of samples are closer to $\mathbf{z}_{cnt}$. See Figure \ref{fig:visual_ts_complete} for the complete figure. By observing these examples, we may extract semantic information that may explain the difference between the abnormal and normal time series. 
\begin{itemize}
    \item Anomalous traffic time series may carry an unusually high amount of traffic data compared with normal traffic time series, as given in the first two examples.
    \item Abnormal traffic time series may bear long and deep sleeping modes in which no traffic is transmitted.
\end{itemize}
Please notice that such semantic information extracted from these examples may be used to identify other anomalous time series as well.  

\subsubsection{Hyperparameter Sensitivity Analysis}
\begin{figure}[!tbp]
\centering
    \subfloat[Effect of $N_E$, $N_L$ and $K$]{
        \begin{minipage}{0.47\columnwidth}
        \includegraphics[width=\columnwidth]{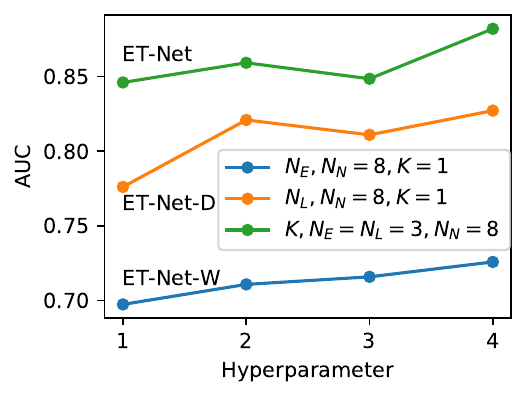}
        \end{minipage}}
    \subfloat[Effect of $N_N$]{
        \begin{minipage}{0.47\columnwidth}
        \includegraphics[width=\columnwidth]{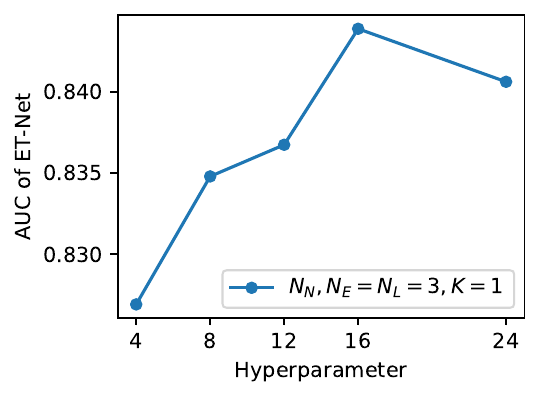}
        \end{minipage}}
\caption{Effect of hyperparameters}
\label{fig:exp_hyper}
\end{figure}

In Table \ref{table:hyper_params}, we list four hyperparameters of ET-Net. $N_E$ determines the number of tasks in the W compression network, while $N_L$ specifies the number of resolution levels used to parse the time series in the D compression network. To evaluate their impact, we vary $N_E$ and $N_L$ from 1 to 4 and report the average AUC of ET-Net-W and ET-Net-D after five runs on the UNSW-IoT dataset, as depicted in Figure \ref{fig:exp_hyper}(a). Our findings suggest that increasing both $N_E$ and $N_L$ enhances performance.
Datasets with heterogeneous temporal dynamics, such as UNSW-IoT, can benefit from an increased number of Gaussian components $K$. This adjustment better estimates the complex distribution of the latent space, as indicated by the green line in Figure \ref{fig:exp_hyper}(a).
Moreover, we investigated the effect of the number of neurons $N_N$ on ET-Net's performance. Figure \ref{fig:exp_hyper}(b) highlights that increasing the number of neurons leads to improved performance.

It is worth noting that the computational complexity of the model depends on the hyperparameters $N_L$, $N_E$, and $N_N$, which have complexities of $O(N_L)$, $O(N_E)$, and $O(N_N^2)$, respectively.  In anomaly detection tasks, selecting the appropriate number of Gaussian components $K$ is crucial, and should be determined based on the complexity of the distribution of the normal samples. For small datasets, it is recommended to appropriately reduce these parameters to avoid overfitting. As demonstrated in Figure \ref{fig:exp_hyper}(b), increasing the number of neurons to 24 results in a decrease in performance. On the other hand, in clustering tasks, the number of GMM components is set equal to the number of clusters.

\subsection{Time Series Clustering Results on Real-world Datasets}
\textbf{Implementation Details:}
We conduct clustering experiments on three real datasets, including UNSW-IoT, IoT23 and cell traffic. Hyperparameters used in the experiment are listed in Table \ref{table:hyper_params}.

\textbf{Effectiveness:}
Table \ref{table:result_cluster} lists the clustering performance of ET-Net and other state-of-the-art methods on three real-world datasets. Two other methods have also been considered for comparison, including
1) AE+K-means in which the clustering is carried out over the latent space representations, which is obtained through the sequence autoencoder.
2) ET-Net+K-means in which the clustering is carried out over the vector embeddings obtained by the W and D compression network.
For both approaches, the same network hyperparameters as ET-Net
are adopted.

The results show that ET-Net outperforms all other state-of-the-art methods in two out of three datasets. This substantiates the effectiveness of the ET-Net for clustering. 
\textcolor{black}{
However, as noted in \cite{shirkhorshidi2015comparison}, there is no universally suitable similarity metric for all tasks. SPIRAL, which utilizes the inner product of representations as a similarity metric, outperforms other distance-based methods in the UNSW-IoT dataset. Moreover, the proposed method performs best among all distance-based methods.}

\begin{table}[!tbp]
\centering
\caption{Clustering performance measured by NMI. Optimal and suboptimal results are bolded and underlined, respectively.}
\label{table:result_cluster}
\scalebox{1}{
    \begin{tabular}{@{}lcccc@{}}
    \toprule
    Method         & UNSW-IoT        & IoT23           & Cell traffic    \\ \midrule
    K-means        & 0.0202          & 0.0399          & 0.0312          \\
    GMM            & 0.0000          & 0.0000          & 0.0107          \\
    K-means + DTW  & 0.5882          & 0.4860          & 0.0189          \\
    K-means + EDR  & 0.5046          & 0.3260          & 0.0318          \\ \midrule
    K-shape        & 0.6071          & 0.4558          & 0.0258          \\
    SPIRAL         & \textbf{0.9138} & 0.4390          & 0.0336          \\
    Autowarp       & 0.1002          & 0.1987          & 0.0281               \\
    DEC            & 0.0202          & 0.3091          & 0.0102          \\
    IDEC           & 0.0201          & 0.2569          & 0.0195          \\
    DTC            & 0.6117          & 0.1862          & 0.0000          \\ \midrule
    AE + K-means   & 0.7033          & 0.6268          & 0.0269          \\
    ET-Net + K-means & 0.6701          & 0.5659          & 0.0366          \\ \midrule
    ET-Net-W         & 0.4023          & \underline{0.6327}    & \underline{0.0542}    \\
    ET-Net-D         & 0.3734          & 0.2694          & 0.0193          \\
    \textbf{ET-Net}           & \underline{0.8304}    & \textbf{0.6753} & \textbf{0.0582} \\ \bottomrule
    \end{tabular}}
\end{table}

\section{Conclusion}
In this paper, we present ET-Net, an unsupervised deep learning approach that learns similarity metrics on event-triggered time series. Through extensive qualitative and quantitative studies, it is revealed that the proposed model can effectively capture the temporal dynamics of event-triggered time series. In addition, a single ET-Net model can be applied to time series with different time granularity with little performance degradation, which shows its robustness.

% ---------------------------------------------
\appendices
% \section*{Appendix}

% % use section* for acknowledgment
% \ifCLASSOPTIONcompsoc
%   % The Computer Society usually uses the plural form
%   \section*{Acknowledgments}
% \else
%   % regular IEEE prefers the singular form
% \fi

% \textcolor{black}{This work was supported by ...}

\ifCLASSOPTIONcaptionsoff
  \newpage
\fi

\bibliographystyle{IEEEtran}
\bibliography{IEEEabrv,ijcai20}


\begin{IEEEbiography}[{\includegraphics[width=1in,height=1.25in,clip,keepaspectratio]{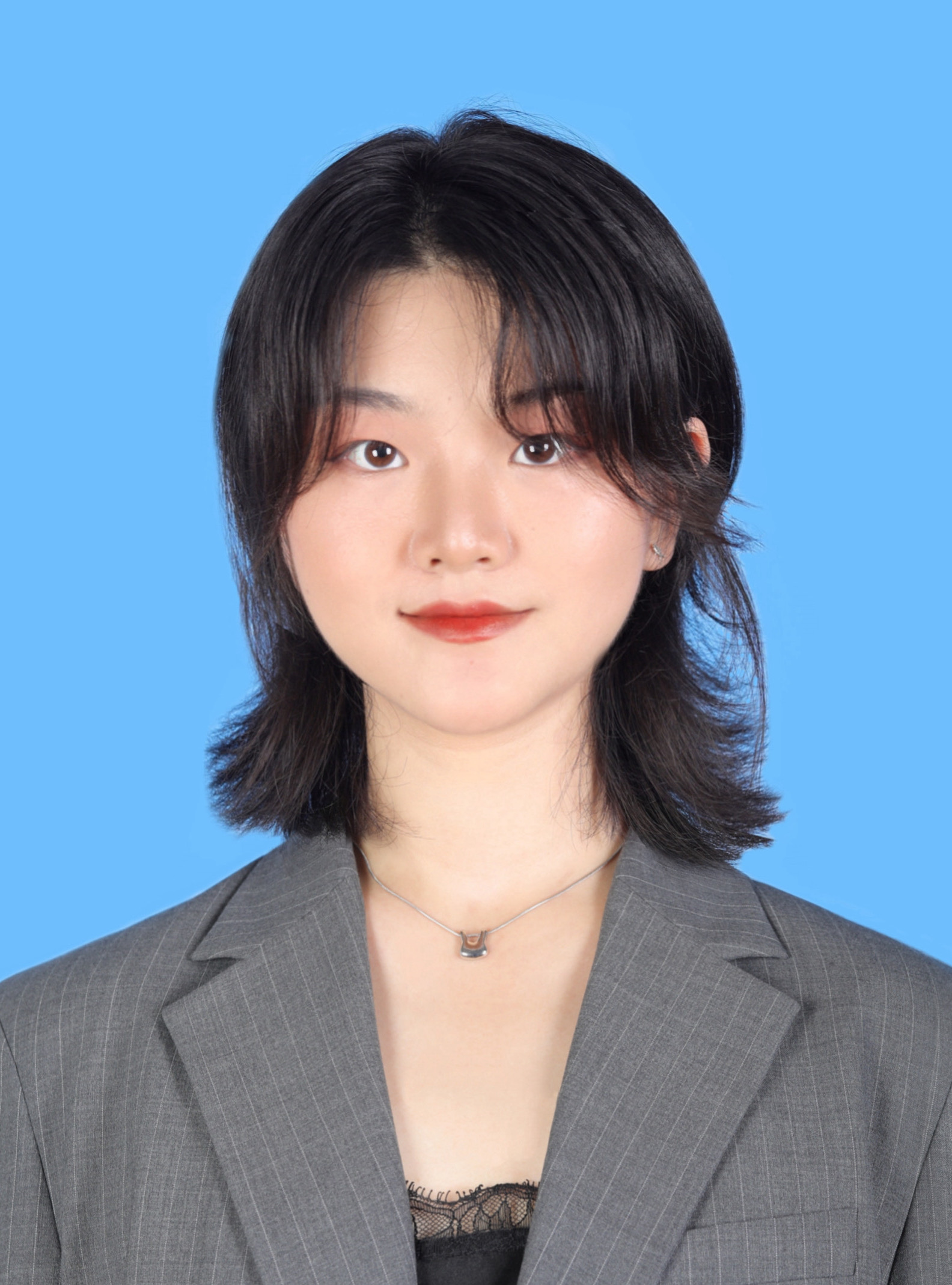}}]
{Shaoyu Dou} received the B.Eng. degree from Hohai University, Nanjing, China, in 2018, and subsequently received the Ph.D. degree from Tongji University, Shanghai, China. She currently holds the position of senior research \& development engineer at Ant Group. Her primary research interests include large language models, AI for IT Operations, big data analytics, and machine learning.
\end{IEEEbiography}

\begin{IEEEbiography}[{\includegraphics[width=1in,height=1.25in,clip,keepaspectratio]{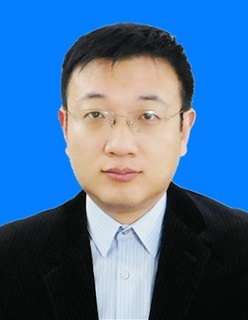}}]
{Kai Yang} (SM'18) received the B.Eng. degree from Southeast University, Nanjing, China, the M.S. degree from the National University of Singapore, Singapore, and the Ph.D. degree from Columbia University, New York, NY, USA.

He is a Distinguished Professor with Tongji University, Shanghai, China. He was a Technical Staff Member with Bell Laboratories, Murray Hill, NJ, USA. He has also been an Adjunct Faculty Member with Columbia University since 2011. He holds over 20 patents and has been published extensively in leading IEEE journals and conferences. His current research interests include big data analytics, machine learning, wireless communications, and signal processing.
\end{IEEEbiography}

\begin{IEEEbiography}[{\includegraphics[width=1in,height=1.25in,clip,keepaspectratio]{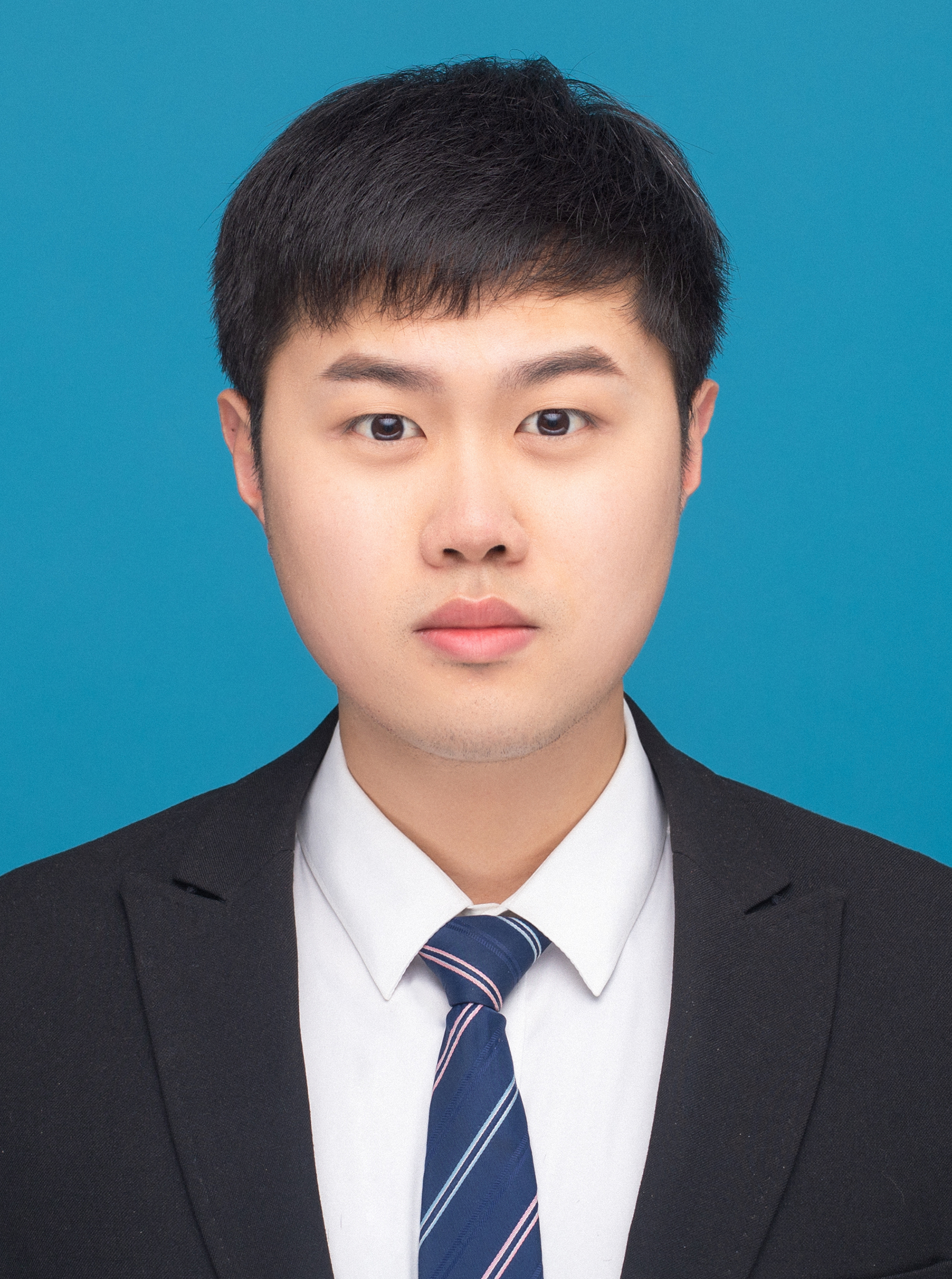}}]
{Yang Jiao} received the B.S. degree from Central South University, Changsha, China, in 2020. He is currently working toward the Ph.D. degree in computer science with the Department of Computer Science and Technology, Tongji University, Shanghai, China. He has authored at top-tier artificial intelligence conferences, such as NeurIPS, ICLR in his research areas, which include machine learning, robust optimization and distributed optimization.
\end{IEEEbiography}

\begin{IEEEbiography}[{\includegraphics[width=1in,height=1.25in,clip,keepaspectratio]{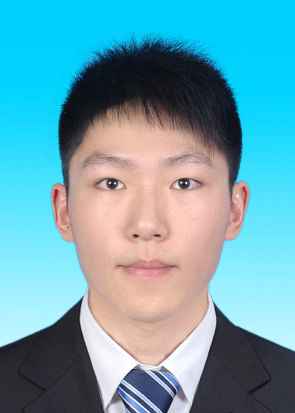}}]
{Chengbo Qiu} received his M.S degree from Huazhong University of Science and Technology, Wuhan, China, in 2018. He is currently pursuing the Ph.D. degree in computer science from the Department of Computer Science, Tongji University, Shanghai, China. His major research interests include big data analytics and machine learning.
\end{IEEEbiography}

\begin{IEEEbiography}[{\includegraphics[width=1in,height=1.25in,clip,keepaspectratio]{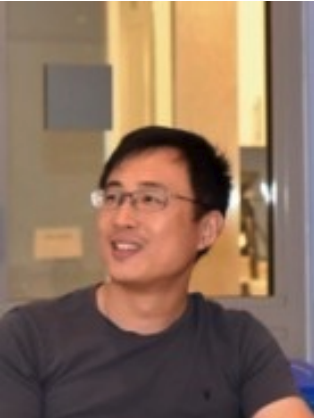}}]
{Kui Ren} received degrees from three different majors, i.e., his Ph.D. in Electrical and Computer Engineering from Worcester Polytechnic Institute, USA, in 2007, M.Eng in Materials Engineering in 2001, and B.Eng in Chemical Engineering in 1998, both from Zhejiang University, China. Professor Kui Ren, AAAS, ACM, CCF, and IEEE Fellow, is currently the dean of the College of Computer Science and Technology at Zhejiang University. He is mainly engaged in research in data security and privacy protection, AI security, and security in intelligent devices and vehicular networks.
\end{IEEEbiography}

%\newpage


%\vfill


\end{document}